%% file: main.tex
\documentclass[10pt,twocolumn,letterpaper]{article}

\usepackage{cvpr}              % To produce the CAMERA-READY version

\usepackage{algorithm}
\usepackage{algpseudocode}
\usepackage{booktabs}
\usepackage{amsfonts}
\usepackage{nicefrac}
\usepackage{multirow}
\usepackage{amsmath}
\usepackage{graphicx}
\usepackage{makecell}
\usepackage{color}         % colors
\usepackage{xcolor}
\usepackage{colortbl}
\usepackage{nicefrac}
\usepackage{diagbox}
\usepackage{pifont}

\usepackage{wrapfig}
\usepackage{enumitem}

\definecolor{LightPurple}{rgb}{0.88,0.88,1}
\definecolor{baselinecolor}{gray}{.9}
\newcommand{\lightgray}[1]{\cellcolor{baselinecolor}{#1}}

\definecolor{cvprblue}{rgb}{0.21,0.49,0.74}
\usepackage[pagebackref,breaklinks,colorlinks,allcolors=cvprblue]{hyperref}

\def\paperID{} % *** Enter the Paper ID here
\def\confName{CVPR}
\def\confYear{2026}

\title{PA-Attack: Guiding Gray-Box Attacks on LVLM Vision Encoders with Prototypes and Attention}

\author{
    Hefei Mei\textsuperscript{1} \quad 
    Zirui Wang\textsuperscript{1} \quad 
    Chang Xu\textsuperscript{2} \quad 
    Jianyuan Guo\textsuperscript{1} \quad 
    Minjing Dong\textsuperscript{1}\thanks{Corresponding author} \\[1ex]
    \textsuperscript{1}City University of Hong Kong \qquad 
    \textsuperscript{2}The University of Sydney \\ 
    {\tt\small \{hefeimei2-c, zrwang23-c\}@my.cityu.edu.hk} \\
    {\tt\small c.xu@sydney.edu.au, \{jianyguo, minjdong\}@cityu.edu.hk}
}

\begin{document}
\maketitle
\input{sec/0_abstract}    
\input{sec/1_intro}
\input{sec/2_method}
\input{sec/3_experiments}

{
    \small
    \bibliographystyle{ieeenat_fullname}
    \bibliography{main}
}

\input{sec/X_suppl}


\end{document}

%% file: sec/0_abstract.tex
% 1. Prototype-Attention-Directed有点怪，你的attention是token，和prototype是两个contribution，全部混一块不理解什么意思。 Potential method names: Prototype-Anchored Attack with attention； Token-Attentive Prototype Attack, Generalizable graybox attack;
% Attentive Prototype-Anchored Attack
% 2. 题目里面尽可能避免shadows，确定完method名之后，你把abstract/intro过遍gpt，让它生成点名字自己组合下
\begin{abstract}
Large Vision–Language Models (LVLMs) are foundational to modern multimodal applications, yet their susceptibility to adversarial attacks remains a critical concern. 
Prior white-box attacks rarely generalize across tasks, and black-box methods depend on expensive transfer, which limits efficiency. The vision encoder, standardized and often shared across LVLMs, provides a stable gray-box pivot with strong cross-model transfer. Building on this premise, we introduce \textbf{PA-Attack} (Prototype-Anchored Attentive Attack).
% To harness the full potential of vision encoders for efficient and generalizable disruption of LVLM, we propose \textbf{PAD-Attack} (Prototype-Attention-Directed Attack), a gray‑box approach that directly targets vision encoders and efficiently disrupts LVLMs. 
PA-Attack begins with a prototype-anchored guidance that provides a stable attack direction towards a general and dissimilar prototype, tackling the attribute-restricted issue and limited task generalization of vanilla attacks. Building on this, we propose a two-stage attention enhancement mechanism: (i) leverage token‑level attention scores to concentrate perturbations on critical visual tokens, and (ii) adaptively recalibrate attention weights to track the evolving attention during the adversarial process. Extensive experiments across diverse downstream tasks and LVLM architectures show that PA‑Attack achieves an average $75.1\%$ score reduction rate (SRR), demonstrating strong attack effectiveness, efficiency, and task generalization in LVLMs. Code is available at https://github.com/hefeimei06/PA-Attack.

\end{abstract}

%% file: sec/1_intro.tex
\section{Introduction}
\label{sec:intro}

Large Vision–Language Models (LVLMs) integrate visual and linguistic modalities, enabling a new generation of multimodal applications~\cite{liu2023visual, zhu2023minigpt, Qwen-VL, team2023gemini, ye2024mplug}. As LVLMs are increasingly deployed in real-world applications, it is imperative to understand their robustness and security vulnerabilities~\cite{liu2025survey, ye2025survey, zhang2024multitrust}. Adversarial attacks, which induce model failures by adding imperceptible perturbations to inputs, pose a significant threat to LVLMs~\cite{schlarmann2023adversarial, bhagwatkar2024towards, zhang2024anyattack, liu2024pandora, you2026unibreak}. 
% Thus, understanding these vulnerabilities becomes essential for developing reliable and safe models.

\begin{figure}
\centerline{\includegraphics[width=1.0\linewidth]{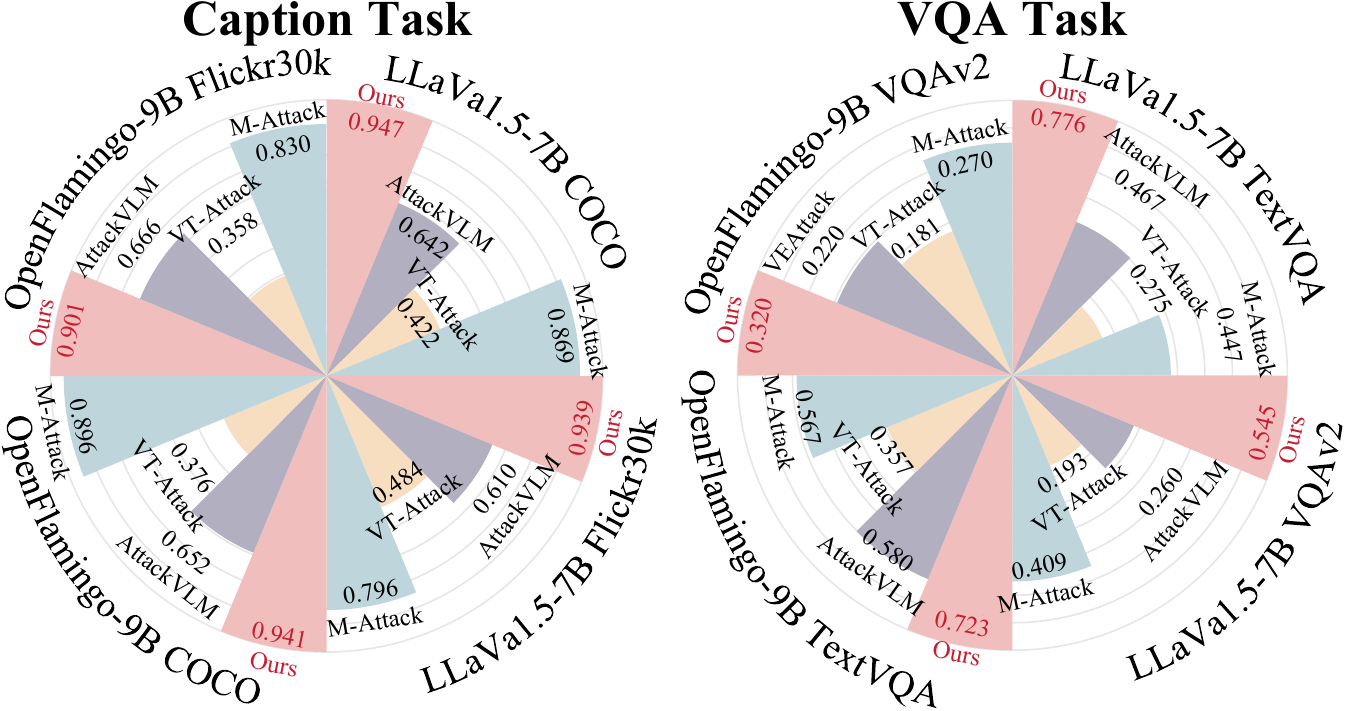}}
    % \vspace{-1em}
	\caption{Adversarial performance on captioning and VQA tasks. The perturbation of black-box M-attack is $\epsilon=\nicefrac{16}{255}$ while that of other gray-box methods is $\epsilon=\nicefrac{2}{255}$.}
    \label{f-introduction}
    \vspace{-1em}
\end{figure}

\begin{figure*}
\centerline{\includegraphics[width=1.0\linewidth]{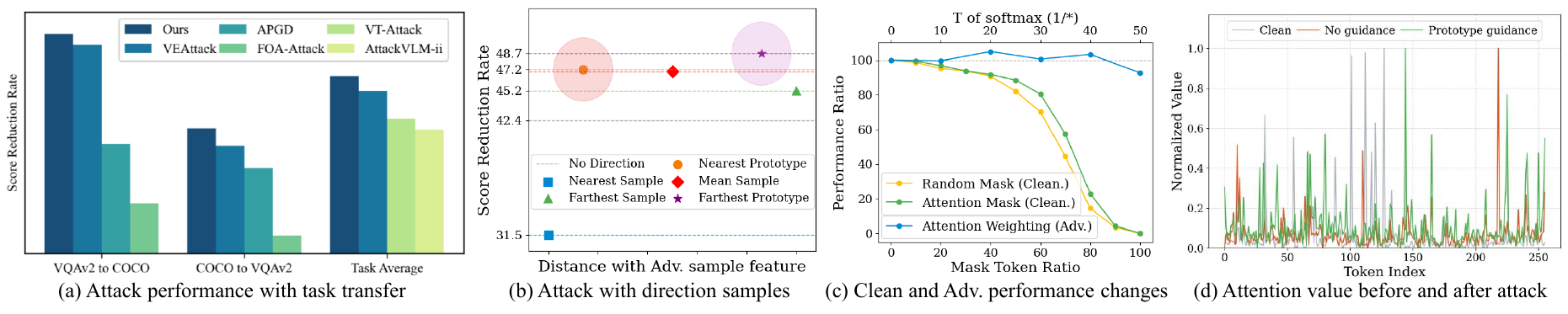}}
	\caption{(a) \textbf{Comparison of task transfer performance score reduction rate (SRR) on LLaVa1.5-7B.} Slash columns compare the task transfer with white-box and black-box attack, while dotted columns represent the multi-task SRR of gray-box. (b) \textbf{Attack SRR w/ and w/o different direction samples.} The red diamond is the baseline without direction samples, and the light orange and purple circles represent their centers as prototypes. (c) \textbf{The ratio of clean and adversarial performance changes.} The blue line has the horizontal axis above (T in softmax) as a variable, and the other lines have the horizontal axis below (the proportion of mask tokens) as a variable. (d) \textbf{Attentions before and after a 50-step attack w/ and w/o prototype guidance.} For comparison, values are normalized with the maximum value.}
    \label{f-evidence}
    \vspace{-1em}
\end{figure*}

Two stubborn challenges for attacking LVLMs lie in efficiency and task generalization~\cite{liu2025survey}, especially as the complexity of large language models (LLMs) and task diversity continue to grow.
Existing white-box attacks~\cite{madry2017towards, croce2020reliable, schlarmann2023adversarial, cui2024robustness} assume full access to massive LVLM parameters but cannot generate adversarial examples that generalize to different LVLM tasks. The strong assumption and low efficiency reduce the attacking practicability.
% Besides their massive parameters, attack overhead is further amplified by their task diversity, particularly for white-box methods that struggle to generalize across tasks~\cite{madry2017towards, croce2020reliable, schlarmann2023adversarial, cui2024robustness}.
Meanwhile, black-box attacks~\cite{zhao2023evaluating, zhang2024anyattack, li2025frustratingly, jia2025adversarial} explored various transfer strategies to achieve high attack transferability among LVLM tasks. However, these strategies always lead to high computational cost and assume large perturbation sizes that reduce stealthiness.
% but were hindered by reliance on complex transfer strategies that consequently lead to low efficiency, while the use of large and semantical perturbations compromise imperceptibility.
As shown in Fig.~\ref{f-introduction}, black-box M-Attack~\cite{li2025frustratingly} struggles to achieve a high score reduction rate (SRR) of LVLM performance, despite employing larger perturbations. 

By contrast, a gray-box attack targeting a foundational LVLM module provides better trade-offs among efficiency, transferability, and practicality since only partial parameters are required and overfitting to a specific LVLM task is relaxed. In particular, attacking the vision encoder offers several advantages. First, most LVLMs, such as LLaVA, Yi-VL, DeepSeek-VL~\cite{liu2023visual,young2024yi,lu2024deepseek}, are built by pairing a shared vision backbone (e.g., CLIP~\cite{radford2021learning}) with different LLMs, \ie, Vicuna, Yi-LLM, DeepSeek-LLM~\cite{chiang2023vicuna,young2024yi,bi2024deepseek}, which makes the vision encoder a common component across diverse LVLMs and an ideal target for attack. Second, the vision encoder typically contains significantly fewer parameters than LLM modules, thereby boosting efficiency. Third, perturbations crafted to disrupt the vision encoder could be more generalizable across LVLM tasks since all downstream multimodal tasks rely on fundamental visual representations.
% A promising solution is to adopt a gray-box setting targeting the accessible vision encoder, as vision encoders (e.g., CLIP~\cite{radford2021learning}) are often open-sourced or standardized and serve as the foundational bottleneck for all multi-modal perception. Consequently, by directly disrupting this foundational component, attacks can efficiently generalize across various downstream tasks, which are heavily dominated by the vision encoder's representations.
% A promising solution is to adopt a gray-box setting targeting a foundational LVLM module, enhancing efficiency and generalization by avoiding transfer strategies and reducing the optimization scope. To this end, the vision encoder is an ideal target, as it is frequently open-sourced and standardized, acting as a common component across diverse LVLMs that couple a shared vision backbone with different, computationally massive LLMs (e.g., LLaVA~\cite{liu2023visual} with Vicuna~\cite{chiang2023vicuna}, Yi-VL~\cite{young2024yi} with Yi-LLM, and DeepSeek-VL~\cite{lu2024deepseek} with DeepSeek-LLM~\cite{bi2024deepseek}, all leveraging CLIP-ViT~\cite{radford2021learning}). Directly disrupting this shared foundational bottleneck is therefore an efficient and highly generalizable attack for various downstream tasks.

However, existing gray-box methods suffer from trade-offs between efficiency and effectiveness. VT-Attack~\cite{wang2024break} requires auxiliary text information from LVLM captions and high attack iterations, leading to significant costs. Meanwhile, AttackVLM-ii~\cite{zhao2023evaluating} simply utilizes cosine similarity supervision targeting the class token, which becomes ineffective. As shown in Fig.~\ref{f-introduction} and Fig.~\ref{f-evidence} (a), with lower attack iterations and perturbations, it is difficult for these gray-box attacks to attack LVLMs in different tasks. 
We mainly attribute it to the attack generalization. Existing gray-box attacks typically maximize the discrepancy between adversarial features and clean ones. However, these attacks always overfit to limited visual attributes without guidance. As shown in the red line of Fig.~\ref{f-evidence} (d), a few specific tokens dominate the attack. Fig.~\ref{f-subject} also shows that the generated adversarial example can hardly generalize to tasks that focus on different visual attributes.
% In essence, existing attacks typically devise objectives to maximize the discrepancy between adversarial features and their clean features. However, the observation in Fig.~\ref{f-evidence} (d) shows that without a proper general guidance, the attack process tends to overfit to a few specific tokens. Fig.~\ref{f-subject} shows that such attack leads to limited visual attributes, hindering the generalization of attacks across downstream tasks.
% , hindering the generalization of attacks across downstream tasks. \textcolor{blue}{Jianyuan: From this sentence, how can reader know from fig2(d) that `limited attack attributes, hindering the generalization attacks across downstream task'} % Hefei: 这里需要和后面的主观图建立一下联系，等有了主观图把这句话完善一下
In addition, Figure~\ref{f-evidence} (c) indicates significant token redundancy in performance impact. Existing approaches treat all tokens uniformly, which expands the search space and wastes effort on redundant tokens, thereby becoming ineffective.
% Jianyuan: 这里并不是not only but also的关系，其实就是同一个问题吧————Hefei：这里想说搜索空间大所以效率低，而存在冗余tokens所以容易有效性低
% \textcolor{blue}{Existing approaches treat all tokens uniformly, expanding the search space, wasting effort on redundant tokens, and ultimately undermining effectiveness.}

In this paper, we propose a novel gray-box attack, PA-Attack (Prototype-Anchored Attentive Attack), to explore the potential of vision encoders for generalized attack in various LVLM tasks. To address the attack generalization issue, we first introduce prototype-anchored guidance to provide a more general attack direction targeting more visual attributes. Simply maximizing an objective without guidance could easily lead to overfitting to a few attributes, limiting the attack's generalization across diverse LVLM tasks. Instead of merely maximizing dissimilarity, we provide a stable directional guide by optimizing the adversarial features towards a pre-computed, highly dissimilar prototype derived from a diverse guidance dataset.
Furthermore, to overcome high-dimensional feature redundancy that hinders effective search for adversarial visual features, we then employ token attention enhancement. This mechanism uses class token attention scores as weights to focus the attack on the most critical patch tokens. Because attention patterns shift during the attack, as shown in Fig.~\ref{f-evidence} (d), we design a two-stage attention refinement framework. This approach recalculates attention weights from the first-stage adversarial sample, adaptively aligning the optimization with the evolving state of the adversarial process to improve disruption.
Extensive experiments demonstrate that PA-Attack achieves superior attack effectiveness with $75.1\%$ score reduction rate (SRR) across diverse downstream tasks and exhibits strong generalization when evaluated on multiple LVLMs. Our results highlight the value of vision‑encoder‑centric attacks and advance gray‑box LVLM attack designs.
%and multimodal model security research.

\section{Related work}

\textbf{Adversarial attacks with full model access on LVLMs.} 
White-box attacks on LVLMs assume full access to model components and typically adapt classic gradient-based methods~\cite{madry2017towards, goodfellow2014explaining, carlini2017towards, croce2020reliable} to optimize inputs with imperceptible budgets. APGD~\cite{schlarmann2023adversarial} shows that imperceptible perturbations can steer captions to misinformation or malicious links even with tiny perturbation budgets. Cui et al.~\cite{cui2024robustness} investigate the robustness of perturbations across prompts, concluding that LVLMs are non robust to visual perturbations and using context in the prompt mitigates failures. UMK~\cite{wang2024white} jointly learns an adversarial image prefix and a text suffix, outperforming uni-modal baselines. Due to their reliance on complete gradient information, the adversarial examples they generate often struggle to achieve generalization across the diverse tasks of LVLMs.

\noindent\textbf{Adversarial attacks with limited model access on LVLMs.} 
Research on generalizable LVLM attacks is increasingly focused on limited-access black-box and gray-box settings. In black-box attacks, AttackVLM~\cite{zhao2023evaluating} establishes an effective transfer strategy that aligns image-to-image features from surrogate models. 
SSA-CWA~\cite{dong2023robust} demonstrates attacks against commercial, closed-source MLLMs like Google's Bard~\cite{bard}.
AnyAttack~\cite{zhang2024anyattack} introduces a self-supervised framework, pre-training a noise generator to create attacks independent of target labels. M-Attack~\cite{li2025frustratingly} proposes an effective baseline using random cropping to embed local semantic details. 
FOA-Attack~\cite{jia2025adversarial} jointly aligns global features with local patch-tokens using clustering and Optimal Transport. 
However, their reliance on complex transfer strategies can reduce efficiency. Gray-box attacks have emerged to balance generalization and efficiency by targeting accessible vision encoders. For instance, MIX.Attack~\cite{tu2023many} misleads LVLMs by aligning with a mix of irrelevant textual concepts in the CLIP embedding space. VT-Attack~\cite{wang2024break} leverages text information to generate more effective adversarial samples. VEAttack~\cite{mei2025veattack} attacks the patch tokens by minimizing their cosine similarity, achieving high efficiency. This work follows the gray-box setting and further explores the potential of the vision encoder in achieving efficient and generalized LVLM attacks.

\begin{figure*}
\centerline{\includegraphics[width=1.0\linewidth]{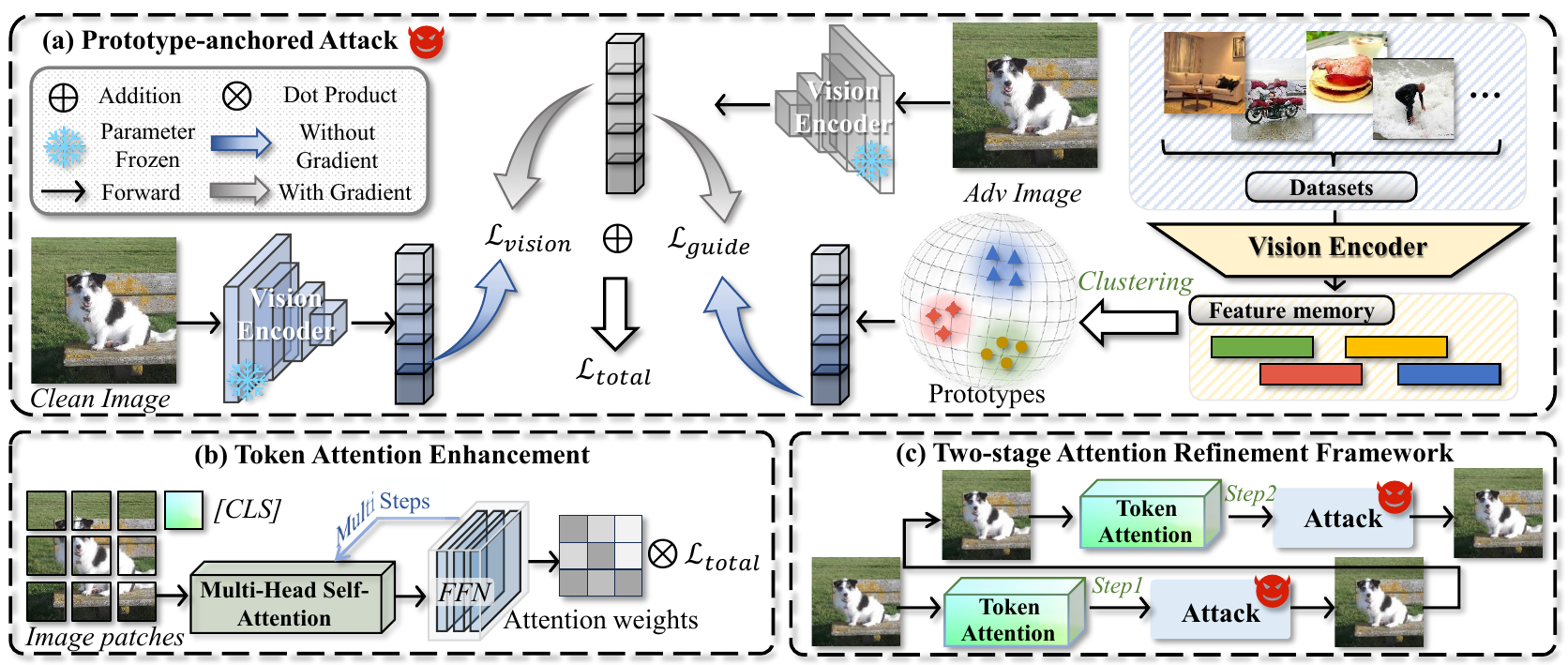}}
	\caption{\textbf{Overview of PA-Attack.} (a) The prototype-anchored guidance includes a vision encoder attack loss and a guidance loss for general degradation across diverse tasks. (b) Attention obtained by averaging the attention of the class token to each patch in Self-Attention across different Heads. (c) The attention weights are adjusted to align with the adversarial image through a two-stage process.}
    \label{f-overview}
    \vspace{-1em}
\end{figure*}

%% file: sec/2_method.tex
\section{Methodology}
\label{method}

To achieve an efficient and generalizable attack on the vision encoder of LVLMs, we introduce PA-Attack, a two-stage optimization framework. As shown in Figure~\ref{f-overview}, we first employ a prototype-anchored objective that drives adversarial features toward a dissimilar prototype, encouraging attacks to cover diverse visual attributes and providing a stable and general attack direction. We further concentrate the limited attacking budget on important tokens by weighting tokens with attention and refreshing these weights with a two-stage framework, where the evolving attention during the adversarial optimization is tracked.

% \subsection{Untargeted visual feature objective}
\subsection{Vision Encoder Attack}
\label{sec:adv_obj}

Let $g$ denote a pre-trained Large Vision-Language Model (LVLM), and let $f$ be its vision encoder. The loss function $\mathcal{L}$ acts as the core attack objective to guide the generation of adversarial samples. Given an evaluated image-text dataset $\mathbb{D}_{eval} \subset \mathcal{X} \times \mathcal{T}$, which contains pairs $(\boldsymbol{x}_i, t_i)$, where $\boldsymbol{x}_i$ is an clean image, and $t_i$ is the constructions for LVLMs. The vision encoder $f: \mathcal{X} \mapsto \mathbb{R}^{N\times d}$ encodes the images to visual tokens $\boldsymbol{v}\in \mathbb{R}^{N\times d}$, which are aligned with text tokens from $\mathcal{T}$. All the tokens are then fed into the LLM to generate the response in an autoregressive manner. During adversarial attacking, let $\mathcal{B}_\epsilon(\boldsymbol{x})=\left\{\boldsymbol{x}^{\prime}:\left\|\boldsymbol{x}^{\prime}-\boldsymbol{x}\right\|_\infty \leq \epsilon\right\}$ denote as an $\ell_\infty$-norm ball for adversarial samples $\boldsymbol{x}^{\prime}$, where $\epsilon$ is a perturbation budget. For $t_q$ in multiple constructions with a single image $\boldsymbol{x}$, the normal optimization problem of white-box can be formulated as
\begin{equation}
    \max _{\boldsymbol{x}+\boldsymbol{\delta}_q\in\mathcal{B}_\epsilon(\boldsymbol{x})} \mathcal{L}\left(g(\boldsymbol{x},t_q), g(\boldsymbol{x}+\boldsymbol{\delta},t_q)\right),
    \label{e-problem-white}
\end{equation}
where the optimized $\boldsymbol{\delta}_q$ denotes an adversarial perturbation for construction $t_q$.
While white-box attacks could disrupt LVLMs successfully, they still struggle with poor task generalization, such as the APGD attack~\cite{croce2020reliable} shown in Fig.~\ref{f-evidence} (a).
Existing black-box attacks~\cite{zhao2023evaluating, li2025frustratingly, jia2025adversarial} explore the generalized attacks, but primarily focus on expensive transfer strategies and rely on large semantical perturbations. 
% As Fig.~\ref{f-evidence} (a) shows, it is difficult for black-box FOA-Attack to be effective under a low perturbation size of $\epsilon=\nicefrac{4}{255}$.
% black-box FOA-Attack remains challenged in effectiveness for the untargeted attack under low perturbation with $\epsilon=\nicefrac{4}{255}$.
Addressing these gaps, we focus on a gray-box setting, targeting only the shared and fundamental vision encoder in LVLMs.
% Let $f$ be a pre-trained vision encoder for LVLMs and $\mathcal{L}$ be the loss function that acts as the core attack objective to guide the generation of adversarial samples. Given an evaluated image-text dataset $\mathbb{D}_{eval} \subset \mathcal{X} \times \mathcal{T}$, which contains pairs $(\boldsymbol{x}_i, t_i)$, where $\boldsymbol{x}_i$ is an clean image, and $t_i$ is the corresponding constructions for LVLMs. 
As the goal of our work is solely attacking the vision encoder $f$ to disrupt the performance of LVLMs, the optimization problem can be reformulated from Eq.~(\ref{e-problem-white}) as
\begin{equation}
    \max _{\boldsymbol{x}+\boldsymbol{\delta}\in\mathcal{B}_\epsilon(\boldsymbol{x})} \mathcal{L}\left(f(\boldsymbol{x}), f(\boldsymbol{x}+\boldsymbol{\delta})\right),
    \label{e-problem}
\end{equation}
% , following the relationship as $\boldsymbol{x}^{\prime}=\boldsymbol{x}+\boldsymbol{\delta}$.
% To efficiently disrupt LVLMs with generalization, we basically transform the attack objective from LVLMs to the vision encoder $f$. 
Following the core optimization goal defined in Eq.~(\ref{e-problem}), we use dissimilarity between the visual features derived from clean images and those perturbed by adversarial perturbations, which can be expressed as
\begin{equation}
    \mathcal{L}_{vision}=-\frac{1}{N}\sum\nolimits_{j}\text{cos}
    \left(f(\boldsymbol{x})_j, f(\boldsymbol{x}+\boldsymbol{\delta})_j\right),
    \label{e-untarget}
\end{equation}
where $j\in\{1,2,\dots,N\}$ is the index of tokens.

\subsection{Prototype-anchored Guidance}
\label{sec:direction}

Our goal is to fully explore the potential of attacking only the vision encoder $f$, which enhances the attack effectiveness and generalization in a gray-box setting. Prior objectives in vision encoder attack (\eg, Eq.~(\ref{e-untarget}) and AttackVLM-ii~\cite{zhao2023evaluating}) primarily maximize the feature discrepancy from clean samples $\boldsymbol{x}$. However, without a proper general guidance, the optimization tends to overfit to a few specific tokens as shown in Fig.~\ref{f-evidence} (d), which concentrate on specific attributes and hinder the generalization across diverse tasks in Fig.~\ref{f-subject}. To address this, we introduce prototype-anchored guidance to the loss function, leading generated adversarial examples to cover diverse visual attributes.

% a prototype-anchored target loss, explicitly providing general guidance for the optimization of attacks.

Given a guidance dataset $\mathbb{D}_{guide}\subset \mathcal{X}$, which does not overlap with the evaluation set as $\pi_1(\mathbb{D}_{eval}) \cap \mathbb{D}_{guide} = \varnothing$, where $\pi_1\left(\mathbb{D}_{eval}\right)=\left\{\boldsymbol{x} \in \mathcal{X} \mid \exists t \in \mathcal{T},(\boldsymbol{x}, t) \in \mathbb{D}_{eval}\right\}$ is the set of all images in $\mathbb{D}_{eval}$. For all images in guidance set $\{\boldsymbol{x}^1_{guide}, \boldsymbol{x}^2_{guide}, \cdots,\boldsymbol{x}^m_{guide}\}$, we extract their features using the vision encoder and save them into a feature memory $\mathcal{M}=\{\boldsymbol{v}^1_{guide}, \boldsymbol{v}^2_{guide}, \cdots,\boldsymbol{v}^m_{guide}\}$ as shown in Fig.~\ref{f-overview}, where $m$ is the number of images in guidance dataset $\mathbb{D}_{guide}$. 
As the visual space $\mathbb{R}^{N\times d}$ is typically large, \eg, $256\times1024$ in CLIP-L/14 with $224\times224$ inputs, the features will retain the top-$m$ principal components using Principal Component Analysis~\cite{abdi2010principal} as $\text{PCA}(\mathcal{M})\mapsto \mathbb{R}^{w}$.
Subsequently, the $K$-Means clustering~\cite{hartigan1979algorithm} is applied to $\text{PCA}(\mathcal{M})$ to group similar spatial token features into $K$ disjoint clusters.
A connection between feature memory $\mathcal{M}$ and $K$ clusters can be established by mapping each feature to a specific cluster. For all $m$ features in the memory, their corresponding cluster indices can be formulated as $\textbf{I}=[i_1, i_2,\dots,i_m] \in \mathbb{R}^{m}$, where $i_t\in\{1,2,\dots,K\}$ denotes the cluster index of a specific feature $\boldsymbol{v}^t_{guide}$. 
% the output is a cluster index $\textbf{I}=[i_1, i_2,\dots,i_m] \in \mathbb{R}^{m}$, where $i_t\in\{1,2,\dots,K\}$.
Then the features that have the same cluster index can be aggregated to form a prototype that contains diverse visual attributes as
\begin{equation} \label{eq:prototype}
\begin{aligned}
    \mathbf{p}^k=\frac{1}{\left|\mathcal{S}_k\right|} \sum_{t \in \mathcal{S}_k} \boldsymbol{v}^t_{guide}, \text{where }\mathcal{S}_k= \{t \; | \; i_t=k\}.
\end{aligned}
\end{equation}
With Eq. (\ref{eq:prototype}), $K$ different prototypes can be constructed as $\mathcal{P}=\{\mathbf{p}^1,\mathbf{p}^2,\cdots,\mathbf{p}^K\}$.
% \textcolor{red}{Notably, the cluster index vector $\textbf{I}$ is consistently associated with the original feature set $\mathcal{M}$, we convert the assigned index set of original feature from $\textbf{I}$ as $\mathcal{S}_k=\{i_t=k\}, k\in\{1,2,\dots,K\}$, which collects all cluster positions of original features in $\mathcal{M}$.} Finally, we aggregate the original features based on the index sets and get the prototypes $\mathbf{p}_k=\frac{1}{\left|\mathcal{S}_k\right|} \sum_{t \in \mathcal{S}_k} \boldsymbol{v}_{guide}$, while $\mathcal{P}=\{\mathbf{p}_1,\mathbf{p}_2,\cdots,\mathbf{p}_K\}$ contains all prototypes of original guidance image features.
In Fig.~\ref{f-evidence} (b), we demonstrate that different guidance samples have different effects on vision encoder attacks, where more distant features could enhance the attack effect to varying degrees. Among them, the furthest prototype achieves the best attack effect by generating more general guidance. Formally, given an input sample $\boldsymbol{x}$ with its feature $\boldsymbol{v}=f(\boldsymbol{x})$, we compute the cosine similarity between $\boldsymbol{v}$ and each $\mathbf{p}^k$ in $\mathcal{P}$ to measure the distance between prototypes and feature $\boldsymbol{v}$. The prototype with the minimum cosine similarity is selected as
\begin{equation}
    k^*=\arg \min _{k \in\{1,2, \ldots, K\}} \cos \left(\boldsymbol{v}, \mathbf{p}^k\right),
    \label{e-index}
\end{equation}
where $k^*$ denotes the index of the prototype selected for attacking $\boldsymbol{v}$. To guide the attack toward maximizing the dissimilarity, we define the prototype-anchored guidance loss as the cosine similarity between $\boldsymbol{v^{\prime}}$ and prototype $\mathbf{p}^{k^*}$ as
% $\mathcal{L}_{guide}=\text{Mean}_j(\text{cos}(\boldsymbol{v^{\prime}}, \mathbf{p}_{k^*}))$. 
\begin{equation}
    \mathcal{L}_{guide}=\frac{1}{N}\sum\nolimits_{j}\text{cos}
    \left(f(\boldsymbol{x})_j, \mathbf{p}^{k^*}_{j}\right).
    \label{e-target}
\end{equation}
Here, the subscript $j$ in $\mathbf{p}^{k^*}_{j}$ serves to index and select the feature corresponding to the $j$-th token within the prototype $\mathbf{p}^{k^*}$. Through integrating the vision encoder attack loss in Eq.~(\ref{e-untarget}) and prototype-anchored guidance loss in Eq.~(\ref{e-target}), the loss function can then be reformulated as
\begin{equation}
% \displaystyle
% \scalebox{0.97}{$
    \mathcal{L}_{total}=\frac{1}{N}\sum\nolimits_{j}[-\text{cos}(\boldsymbol{v}_j, \boldsymbol{v^{\prime}}_j)+\lambda \cdot \text{cos}(\boldsymbol{v^{\prime}}_j,\mathbf{p}^{k^*}_{j})],
% $}    
\end{equation}
where $\boldsymbol{v}=f(\boldsymbol{x})$, $\boldsymbol{v^{\prime}}=f(\boldsymbol{x}+\boldsymbol{\delta})$ are visual features of clean and adversarial images, and $\lambda$ is a hyperparameter that balances the loss terms.

\subsection{Token Attention Enhancement}
\label{sec:attention}

While the prototype-anchored guidance provides a more reliable optimization path, it still suffers from the high dimensionality and structural redundancy of visual features $\boldsymbol{v}\in \mathbb{R}^{N\times d}$, which makes the perturbation optimization difficult and inefficient. As shown in Fig.~\ref{f-evidence} (c), when evaluating token mask proportion versus LVLM performance, the model retains substantial functionality even at a $50\%$ mask proportion, indicating that most tokens are redundant and each token makes an unequal contribution to downstream tasks. Since the maximum budget is limited, we need to focus on more important features during the attack rather than wasting effort on less critical token features to guarantee attack generalization. Observing that preserving higher-attention tokens (via Attention Mask) maintains better performance than random preservation, we use attention scores as weights to prioritize critical tokens in perturbation. 
% This focus enhances attack effectiveness by targeting core visual information.

A vision encoder $f$ splits the input image into $n$ patches and concatenates a class token to form the initial token sequence. This token sequence is subsequently processed through a stack of $L$ layers, each consisting of a multi-head self-attention (MHSA) module and a feed-forward network (FFN). For the MHSA of $l$-th layer and $i$-th head, the corresponding key and features could be $\boldsymbol{K}^l_i, \boldsymbol{V}^l_i\in\mathbb{R}^{(N+1)\times d}$. The class token feature $\boldsymbol{y}^l_i$ is calculated by
\begin{equation}
    \boldsymbol{y}_i^l=\boldsymbol{a}_i^l \boldsymbol{V}_i^l=\operatorname{softmax}\left(\frac{\boldsymbol{q}_i^l \boldsymbol{K}_i^{l T}}{\sqrt{d}}\right) \boldsymbol{V}_i^l,
\end{equation}
where $\boldsymbol{q}_i^l$ is the query feature of the class token. As the class token aggregates global image information, its attention to patch tokens serves as a reliable indicator of patch contribution~\cite{liang2022not}. To obtain a single attention weight for each patch token, we average the attention scores between the class token and each patch token across all $h$ heads as $\boldsymbol{a}^l=\frac{1}{H}\sum_{i} \boldsymbol{a}_i^l$. So the final weights can be denoted as
\begin{equation}
    \boldsymbol{w}_j=\operatorname{softmax}(\boldsymbol{a}^l)=\frac{e^{\boldsymbol{a}^{l}_j/T}}{\sum^n_{j=1}e^{\boldsymbol{a}^{l}_j/T}},
    \label{e-attention}
\end{equation}
where $T$ is a temperature parameter for softmax, and we show its effects in Fig.~\ref{f-evidence} (c). We choose the middle layer as layer $l$ for the computation. Finally, these attention weights $\boldsymbol{w}=\{\boldsymbol{w}_1, \boldsymbol{w}_2,\cdots,\boldsymbol{w}_N\}$ are incorporated into the optimization objective, and the loss can be formulated as
\begin{equation}
\scalebox{0.92}{$ \displaystyle
    \mathcal{L}=-\frac{1}{N}\sum\nolimits_{j}\boldsymbol{w}_j\cdot[-\text{cos}(\boldsymbol{v}_j, \boldsymbol{v^{\prime}}_j)+\lambda \cdot \text{cos}(\boldsymbol{v^{\prime}}_j,\mathbf{p}^{k^*}_{j})].
$}
    \label{e-finalL}
\end{equation}
% During the attack process, both attention scores $\boldsymbol{a}^l$ and token values $\boldsymbol{v}$ are dynamic and change as illustrated in Fig.~\ref{f-evidence} (d). 
During the attack process, the attention scores $\boldsymbol{a}^l$ exhibit dynamic behavior as illustrated in Fig.~\ref{f-evidence} (d), where the attention map of the adversarial image $\boldsymbol{a}^l(\boldsymbol{x^\prime})$ (green distribution) shows a shift compared to that of the clean image $\boldsymbol{a}^l(\boldsymbol{x})$ (gray distribution). This shift guides the model to gradually focus on less robust features, such as background elements or distracting objects. To dynamically identify and prioritize the optimization for tokens truly critical to the adversarial process, we design a two-stage attention refinement framework, which adjusts attention weights $\boldsymbol{w}$ based on the evolutionary state of adversarial samples $\boldsymbol{x}^\prime$.

Specifically, we first extract the token weights $\boldsymbol{w}_{s1}$ from the attention $\boldsymbol{a}^l(\boldsymbol{x})$ during the forward propagation of the clean image $\boldsymbol{x}$ through the vision encoder. Then, we take Eq.~(\ref{e-finalL}) as the optimization objective $\mathcal{L}(f(\boldsymbol{x}),f(\boldsymbol{x^\prime}),\boldsymbol{w}_{s1}, \mathbf{p}^{k^*})$ in Eq.~(\ref{e-finalL}) to prioritize perturbing and perform $S_1$ attack steps in the first stage. Therefore, the optimization of the first stage can be formulated as
\begin{equation}
\scalebox{0.9}{$ \displaystyle
\boldsymbol{x^\prime}_{i+1} \leftarrow \boldsymbol{x^\prime}_{i} + \alpha \cdot \text{sign}(\nabla_{\boldsymbol{x^\prime}_{i}} \mathcal{L}(f(\boldsymbol{x}),f(\boldsymbol{x^\prime}_i),\boldsymbol{w}_{s1}, \mathbf{p}^{k^*})),
$}
\label{e-stagei}
\end{equation}
where $i$ is from $0$ to $S_1-1$, serving as each step in the first-stage attack, and $\alpha$ is the step size. After generating the adversarial image $\boldsymbol{x^{\prime}}_{s1-1}$, the second stage focuses on refining the attention weights to the adversarial process. We re-input $\boldsymbol{x^{\prime}}_{s1-1}$ into the vision encoder and recalculate the attention weights $\boldsymbol{w}_{s2}$ in the second stage using the attention $\boldsymbol{a}^l(\boldsymbol{x^\prime}_{s1-1})$ and Eq.~(\ref{e-attention}). Then we reuse Eq.~(\ref{e-finalL}) as the objective function $\mathcal{L}(f(\boldsymbol{x}),f(\boldsymbol{x^\prime}),\boldsymbol{w}_{s2}, \mathbf{p}^{k^*})$ in Eq.~(\ref{e-finalL}) to perform another $S_2$ attack steps, which can be denoted as
\begin{equation}
\scalebox{0.88}{$ \displaystyle
\boldsymbol{x^\prime}_{r+1} \leftarrow \boldsymbol{x^\prime}_{r} + \alpha \cdot \text{sign}(\nabla_{\boldsymbol{x^\prime}_{r}} \mathcal{L}(f(\boldsymbol{x}),f(\boldsymbol{x^\prime}_r),\boldsymbol{w}_{s2}, \mathbf{p}^{k^*})).
$}
\label{e-stagej}
\end{equation}
The attack step $r$ is from $S_1$ to $S_{1}+S_{2}-1$, and final adversarial sample in the second stage could be $\boldsymbol{x^{\prime}}_{S1+S2-1}$. The attacking process is concluded in Alg.~\ref{alg:LAA}. And more attention results are presented in the supplementary materials.

% Specifically, we first extract the token attention weights $\boldsymbol{w}_{s1}$ through Eq.~(\ref{e-attention}) from clean image $\boldsymbol{x}$, then calculate the prototype index $k^*$ and take Eq.~(\ref{e-finalL}) as the optimization objective $\mathcal{L}(f(\boldsymbol{x}),f(\boldsymbol{x^\prime}_i),\boldsymbol{w}_{s1}, \mathbf{p}_{k^*})$ in Eq.~(\ref{e-finalL}) to prioritize perturbing. We perform $S_1$ attack steps in the first stage and obtain the first-stage adversarial sample $\boldsymbol{x^{\prime}}_{s1}$. The second stage focuses on refining the attention weights to the adversarial process. We re-input $\boldsymbol{x^{\prime}}_{s1}$ into the vision encoder and recalculated the attention weights $\boldsymbol{w}_{s2}$ in the second stage using Eq.~(\ref{e-attention}) and then reuse Eq.~(\ref{e-finalL}) as the objective function $\mathcal{L}(f(\boldsymbol{x}),f(\boldsymbol{x^\prime}_i),\boldsymbol{w}_{s2}, \mathbf{p}_{k^*})$ in Eq.~(\ref{e-finalL}) to perform another $S_2$ attack steps, generating the final adversarial sample $\boldsymbol{x^{\prime}}$. The attacking process is concluded in Alg.~\ref{alg:LAA}.

\begin{algorithm}[t]
\caption{PA-Attacking Procedure}
\label{alg:LAA}
\begin{algorithmic}[1] 
\Require clean image $\boldsymbol{x}$, perturbation budget $\epsilon$, stage-one iterations $S_1$, stage-two iterations $S_2$, loss function $\mathcal{L}$, prototype $\mathcal{P}$ from guidance dataset $\mathbb{D}_{guide}$, step size $\alpha$, random initialization range $\boldsymbol{\eta} \in [0, \epsilon]$.
\State $\boldsymbol{x^{\prime}}_0 \leftarrow \boldsymbol{x} + \text{Uniform}(-\boldsymbol{\eta}, \boldsymbol{\eta})$ \Comment{Random start}
\State $k^*=\arg \min _k\ \cos \left(\boldsymbol{v}, \mathbf{p}^k\right), \mathbf{p}^k\in\mathcal{P}$
\Comment{Index in Eq.~(\ref{e-index})}
\State $\boldsymbol{w}_{s1}=\text{softmax}(\boldsymbol{a}^l\leftarrow f(\boldsymbol{x}))$
\Comment{Token weights Eq.~(\ref{e-attention})}
\For{$i = 0$ to $S_1 - 1$}  
    \State Compute $\mathcal{L}(f(\boldsymbol{x}),f(\boldsymbol{x^\prime}_i),\boldsymbol{w}_{s1}, \mathbf{p}^{k^*})$ in Eq.~(\ref{e-finalL})
    \State $\boldsymbol{x^\prime}_{i+1} \leftarrow \boldsymbol{x^\prime}_{i} + \alpha \cdot \text{sign}(\nabla_{\boldsymbol{x^\prime}_{i}} \mathcal{L})$
    \Comment{Eq.~(\ref{e-stagei})}
    \State $\boldsymbol{x^\prime}_{i+1} \leftarrow \text{clip}(\boldsymbol{x^\prime}_{i+1}, \boldsymbol{x} - \epsilon, \boldsymbol{x} + \epsilon)$
    \Comment{To $\ell_\infty$-ball}
\EndFor
\State $\boldsymbol{w}_{s2}=\text{softmax}(\boldsymbol{a}^l\leftarrow f(\boldsymbol{x^\prime}_{S_1-1}))$
\Comment{Token weights}
\For{$r = S_1$ to $S_1 + S_2 - 1$}  
    \State Compute $\mathcal{L}(f(\boldsymbol{x}),f(\boldsymbol{x^\prime}_r),\boldsymbol{w}_{s2}, \mathbf{p}^{k^*})$ in Eq.~(\ref{e-finalL})
    \State $\boldsymbol{x^\prime}_{r+1} \leftarrow \boldsymbol{x^\prime}_{r} + \alpha \cdot \text{sign}(\nabla_{\boldsymbol{x^\prime}_{r}} \mathcal{L})$
    \Comment{Eq.~(\ref{e-stagej})}
    \State $\boldsymbol{x^\prime}_{r+1} \leftarrow \text{clip}(\boldsymbol{x^\prime}_{r+1}, \boldsymbol{x} - \epsilon, \boldsymbol{x} + \epsilon)$
    \Comment{To $\ell_\infty$-ball}
\EndFor \\
\Return $\boldsymbol{x}_{\text{adv}}$; \Comment{$\boldsymbol{x^\prime}_{S_1+S_2-1} \rightarrow \boldsymbol{x}_{\text{adv}}$}
\end{algorithmic}
\end{algorithm}

%% file: sec/3_experiments.tex
\section{Experiments}

\begin{table*}
\caption{Comparison of PA-Attack with different gray-box attacks across different LVLMs and tasks. For each dataset, we present the performance ($\downarrow$) after the gray-box attack, and finally, we calculate the average score reduction rate (SRR) ($\uparrow$) across all datasets.}
\input{Tables/AttackLVLM}
\label{tab:attackvlm}
\vspace{-1em}
\end{table*}

\subsection{Experimental settings}

\textbf{Datasets.} To demonstrate the generalization of our proposed method across diverse downstream tasks of LVLMs, we select three representative tasks, which are image captioning, visual question answering (VQA), and hallucination detection. For the image captioning task, we adopt COCO~\cite{lin2014microsoft} and Flickr30k~\citep{plummer2015flickr30k} datasets for evaluation. For the VQA task, we choose two extensively utilized datasets, namely TextVQA~\cite{singh2019towards} and VQAv2~\cite{goyal2017making}, to evaluate the attack performance. Finally, for the hallucination task, we construct our evaluation by selecting images from the COCO~\cite{lin2014microsoft} dataset and leveraging the annotation labels provided by the POPE~\citep{li2023evaluating}. Following the setting in \cite{mei2025veattack}, we randomly sampled 500 samples in each task for conducting adversarial attacks and evaluating.

\noindent\textbf{Implementation Settings.} In prototype-anchored guidance, we randomly select the guidance images from the COCO dataset. The number of images in the guidance dataset is set to $m=3000$, and we use Principal Component Analysis (PCA) to retain the top-$w=1024$ principal components of the visual tokens. Subsequently, we apply K-Means clustering to get $K=20$ disjoint prototypes from the guidance of PCA features. The balance parameter of two losses is $\lambda=1.0$. In the attention enhancement, we choose the attention from the middle layer while the temperature for softmax is set to $T=1/20$. The number of attack iterations is set to $S_1=50$ in the first stage, and $S_2=100$ in the second stage. The methods AttackVLM-ii~\cite{zhao2023evaluating} and VEAttack~\cite{mei2025veattack} that we use for comparison set the iterations as $S=100$, and VT-Attack~\cite{wang2024break} set as $S=150$.
Following the setting in VEAttack~\cite{mei2025veattack}, we set the perturbation budges as $\epsilon=2/255$ and $\epsilon=4/255$ to ensure the imperceptible of attacks, and the attack step size is set to $\alpha=1/255$. We evaluate our effectiveness across three LVLMs namely LLaVa1.5-7B~\citep{liu2023visual}, LLaVa1.5-13B~\citep{liu2023visual} and OpenFlamingo-9B (OF-9B)~\citep{awadalla2023openflamingo}. All experiments were conducted on a NVIDIA-A6000 GPU.

\noindent\textbf{Evaluation metrics.} To clearly demonstrate the performance changes of each task, we first show the performance of different tasks after the attack. For image captioning tasks, we use the CIDEr~\citep{vedantam2015cider} score, while for VQA tasks, we measure them by VQA accuracy~\citep{antol2015vqa}. Following POPE~\cite{li2023evaluating}, the hallucination benchmark is evaluated by F1-score. As a gray-box attack, we aim to disrupt the performance of LVLMs more generally and effectively. Therefore, we use the score reduction rate (SRR) to represent the effectiveness of the attack, which can be calculated as:
\begin{equation}
    \text{SRR} = 1 - (\text{Score}_{adv}/\text{Score}_{clean})
\end{equation}
where $\text{Score}_{clean}$ and $\text{Score}_{adv}$ denote the performance before and after the attack. The overall performance across tasks and the ablations is expressed using SRR.

\subsection{Comparison results}

Table~\ref{tab:attackvlm} shows the comparison of our proposed PA-Attack with different gray-box attacks, including MIX.Attack, VT-Attack, AttackVLM-ii, and VEAttack, where AttackVLM-ii follows the black-box attack paradigm targeting vision encoders, treating the accessible vision encoder as a naive baseline. To visually demonstrate performance, we report the corresponding benchmark metrics for each single dataset. The results show large absolute drops across models, such as $115.5$ to $4.1$ in the COCO dataset and LLaVa1.5-7B and $60.1$ to $3.7$ in the Flickr30k dataset and OpenFlamingo-9B. For task generalization, we summarize effectiveness with the score reduction rate (SRR), and our PA-Attack delivers the best average SRR on every model across all tasks. With perturbation budgets as $2/255$ and $4/255$, PA-Attack reaches $77.1\%$ and $79\%$ on LLaVA-1.5-7B, $63.4\%$ and $66.4\%$ on OF-9, which surpasses the strongest gray-box VEAttack by $11.1\%$ and $6.7\%$ in average, and also outperforms the black-box naive baseline AttackVLM-ii by $27.7\%$ and $18.1\%$ in average. Although VT-Attack shows occasional advantages on VQAv2 and LLaVA models, its performance degrades markedly under a small perturbation budget $\epsilon=2/255$ and the advantage does not generalize across tasks, whereas our method achieves a consistently more general and effective attack. Finally, PA-Attack remains highly effective under small perturbations, since even at $\epsilon=2/255$ it already drives all captioning metrics to low single digits and surpasses all models and tasks.

\subsection{Ablation study}

\begin{figure*}
\centerline{\includegraphics[width=1.0\linewidth]{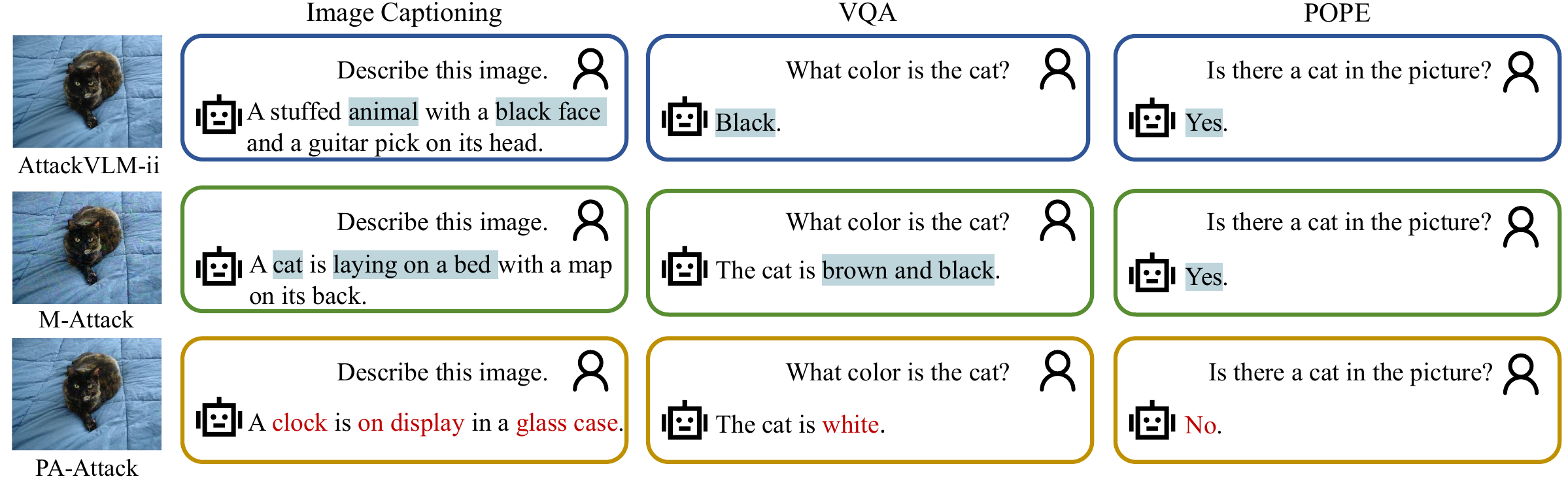}}
    \vspace{-0.3em}
	\caption{\textbf{Comparison of the responses of LLaVa1.5-7B with different attacks.} The attributes with a blue background remain unchanged, while the red texts indicate that the attributes have changed.}
    \label{f-subject}
    \vspace{-0.9em}
\end{figure*}

\begin{table}
\caption{Ablation study on the components of PA-Attack with SRR as the metric, where PG denotes the Prototype-anchored Guidance, AE is the Attention Enhancement, and TS is a two-stage attention refinement framework.}
\input{Tables/module}
\label{tab:module}
\vspace{-1em}
\end{table}

\noindent\textbf{Ablation of each component in PA-Attack.} In Table~\ref{tab:module}, we show the contribution of the prototype-anchored guidance (PG), attention enhancement (AE), and the two-stage refinement (TS) using SRR on LLaVa1.5-7B and four datasets with $\epsilon=4/255$. Starting from the 100-step baseline, introducing PG yields consistent gains, lifting SRR from $93.8\%$ to $95.5\%$ on COCO, and from $72.8\%$ to $76.3\%$ on TextVQA. AE alone produces mixed changes at 100 steps, improving Flickr30k and VQAv2 while leaving COCO and TextVQA essentially unchanged. Combining PG and AE at 100 steps delivers a clear improvement in SRR over the baseline across all datasets, with the largest increase on VQAv2, where SRR rises from $48.4\%$ to $54.2\%$, which proves the effectiveness of the two modules we designed. To ensure a fair comparison of the two-stage framework, we also set the baseline to 150 iterations. Increasing the optimization budget to 150 steps improves the baseline, but algorithmic components remain the dominant factor. With PG and AE at 150 steps, SRR further reaches $96.2\%$ on COCO, $81.1\%$ on TextVQA, which are higher than the 150-step baseline. Introducing the two-stage refinement produces the best results, reaching $96.5\%$ on COCO and $86.3\%$ on TextVQA, which corresponds to improvements over the 150-step baseline and shows the effectiveness of our attention refinement module.

\noindent\textbf{Ablation of loss balance weight.} In Table~\ref{tab:lambda}, we study the balance weight $\lambda$ between the vision encoder attack loss and the prototype-directed term. It can be seen that moderate weight with $\lambda=1.0$ achieves the best overall SRR. A smaller weight $\lambda=0.5$ weakens prototype guidance, which could hurt the QA datasets that have a variety of instructions and may require a more general perturbation. Increasing $\lambda$ to $2.0$ overemphasizes the guidance term, which may affect the degree of damage caused by the vision encoder attack and causes small drops on each dataset. Thus, we adopt the weight $\lambda=1.0$ for all experiments.

\begin{table}
\caption{Ablation study on the weight $\lambda$ of the Prototype-anchored Guidance loss with SRR as the metric.}
\input{Tables/lambda}
\label{tab:lambda}
\end{table}

\begin{table}
\caption{Impact of prototype construction parameters, where $m$ is the number of images in the guidance image dataset, $w$ is the dimension after PCA, and $K$ is the number of prototypes.}
\input{Tables/prototype}
\label{tab:prototype}
\vspace{-1em}
\end{table}

\noindent\textbf{Ablation of the parameters in prototype construction.}
Table~\ref{tab:prototype} presents the impact of the key parameters for prototype construction. We first observe that all parameter variations outperform the baseline, demonstrating the general effectiveness of prototype-anchored guidance. Our method achieves its peak average performance of $74.4\%$ with $m=3000$. Fixing this optimal $m$, we analyze the PCA dimension $w$. Setting $w=1024$ proves optimal, while increasing the dimension to 2048 leads to a significant performance degradation, resulting in a score of $70.7\%$. This indicates that a more compact dimension with $w=1024$ effectively filters out irrelevant variance. Finally, varying the number of prototypes $K$ reveals that $K=20$ yields the highest score, slightly outperforming both fewer clusters $K=10$ and more clusters $K=30$, suggesting $K=20$ provides the best granularity. Therefore, we adopt $m=3000$, $w=1024$, and $K=20$ as our default configuration.

\begin{figure}
\centerline{\includegraphics[width=1.0\linewidth]{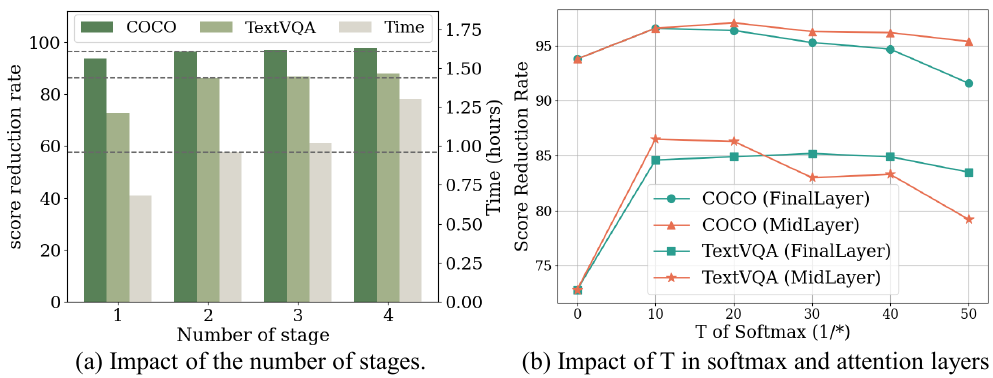}}
	\caption{(a) \textbf{Ablation of the number of stages in adaptively attention refinement.} Slash columns compare the score reduction rate, while dotted columns represent the attack time. (b) \textbf{Ablation of $T$ in softmax and layers $l$ in attention enhancement.}}
    \label{f-stage}
    \vspace{-1em}
\end{figure}

\noindent\textbf{Ablation of the number of stages.} 
We analyze the impact of the number of refinement stages in Fig.~\ref{f-stage} (a). The score reduction rate steadily improves as the number of stages increases, which validates that refining attention allows the attack to progressively focus on more critical features of the adversarial process. However, there is a trade-off between performance and efficiency. Notably, the most significant performance leap is observed when transitioning from one to two stages. In contrast, the gains from subsequent stages become marginal. Therefore, we adopt two stages as our default configuration to achieve an optimal balance between attack effectiveness and efficiency.

\noindent\textbf{Ablation of the $T$ in softmax and layers in attention.} 
As shown in Fig.~\ref{f-stage} (b), we analyze the impact of the source layer for attention extraction and the softmax temperature $T$ in Eq.~(\ref{e-attention}), which is plotted as $1/T$. The performance on both COCO and TextVQA peaks when $T$ is set around $1/20$. This indicates that a large $T$ flattens the distribution, failing to prioritize critical tokens and thus yielding lower scores. Conversely, a small $T$ creates a sharp distribution, which also degrades performance, focusing too narrowly and ignoring other relevant features. Furthermore, at $T=1/20$, using the MidLayer outperforms the FinalLayer. We hypothesize this is because mid-level features retain richer object information, which is beneficial for guiding the attack. Therefore, we adopt the MidLayer for attention extraction and set $T=1/20$ as our default.

\noindent\textbf{Ablation of different guidance samples.} 
We validate the effectiveness of prototype-anchored guidance by comparing it against several guidance samples in Table~\ref{tab:distance}. It can be seen that naively selecting the Nearest sample (the highest cosine similarity) from the guidance set $\mathcal{M}$ will cause the score to significantly decrease. This confirms that optimizing toward a highly similar feature provides a conflicting adversarial signal. Using a Farthest sample or Mean sample yields better results, while the Nearest also shows effective, demonstrating the benefit of using a generalized sample as a stable guidance. Crucially, selecting the Farthest prototype as defined in Eq.~(\ref{e-index}) achieves the highest average score of $82.9\%$, indicating that the guidance is both semantically general and maximally divergent, providing the strongest and most stable optimization for the adversarial process.

\begin{table}
\caption{Impact of different samples for the guidance objective.}
\vspace{-0.5em}
\input{Tables/distance}
\label{tab:distance}
\vspace{-1.5em}
\end{table}

\section{Conclusion}

In this paper, we propose \textbf{PA-Attack}, a novel gray-box adversarial attack that effectively targets the shared vision encoders of Large Vision-Language Models. Our method addresses the critical challenges of attack generalization and efficiency. By introducing a prototype-anchored guidance, PA-Attack establishes a stable and general attack direction, preventing overfitting to limited visual attributes. Furthermore, our two-stage attention enhancement mechanism strategically focuses perturbations on the most critical visual tokens and adaptively calibrates to the evolving attention patterns during the attack. Extensive experiments demonstrate that PA-Attack achieves a state-of-the-art $75.1\%$ average score reduction rate, showing superior effectiveness and strong generalization across diverse downstream tasks and LVLM architectures. Our findings underscore a significant vulnerability of sharing vision backbones, highlighting the urgent need to develop more robust defenses for foundational multi-modal systems.

%% file: Tables/AttackLVLM.tex
\centering
\resizebox{\textwidth}{!}{
    \scriptsize
\tabcolsep=0.15cm
  \begin{tabular}{c|c|cc|cc|cc|cc|cc|cc}
    \toprule
    \multirow{2}*{Model} & Attack & \multicolumn{2}{c|}{COCO $\downarrow$} & \multicolumn{2}{c|}{Flickr30k $\downarrow$} & \multicolumn{2}{c|}{TextVQA $\downarrow$} & \multicolumn{2}{c|}{VQAv2 $\downarrow$} & \multicolumn{2}{c|}{POPE $\downarrow$} & \multicolumn{2}{c}{Average SRR $\uparrow$}\\

     & Perturbation $(\epsilon)$ & $\nicefrac{2}{255}$ & $\nicefrac{4}{255}$ & $\nicefrac{2}{255}$ & $\nicefrac{4}{255}$ & $\nicefrac{2}{255}$ & $\nicefrac{4}{255}$ & $\nicefrac{2}{255}$ & $\nicefrac{4}{255}$ & $\nicefrac{2}{255}$ & $\nicefrac{4}{255}$ & $\nicefrac{2}{255}$ & $\nicefrac{4}{255}$\\

    \midrule

    \multirow{6}*{\rotatebox[origin=c]{90}{{LLaVa1.5-7B}}} & \lightgray Clean & \multicolumn{2}{c|}{\lightgray 115.5} & \multicolumn{2}{c|}{\lightgray 77.5} & \multicolumn{2}{c|}{\lightgray 37.1} & \multicolumn{2}{c}{\lightgray 74.5} & \multicolumn{2}{c}{\lightgray 84.5} & \multicolumn{2}{c}{\lightgray 0.0}\\

     & MIX.Attack~\cite{tu2023many} & 67.5 & 55.4 & 42.1 & 35.9 & 24.6 & 19.1 & 59.4 & 57.9 & 72.0 & 69.1 & 31.2 & 38.9 \\
     
     & VT-Attack~\cite{wang2024break} & 66.8 & 29.5 & 40.0 & 19.9 & 26.9 & 16.1 & 60.1 & \textbf{28.3} & 67.1 & 58.5 & 31.6 & 59.6 \\

     & AttackVLM-ii~\cite{zhao2023evaluating} & 41.3 & 26.3 & 30.2 & 20.3 & 19.7 & 14.0 & 55.1 & 50.4 & 69.0 & 61.6 & 43.3 & 54.5\\

     & VEAttack~\cite{mei2025veattack} & 10.8  & 7.1 & 10.7 & 6.3 & 13.8 & 10.1 & 42.9 & 38.4 & 47.5 & 42.8 & 65.2 & 71.2 \\

     & PA-Attack (ours) & \textbf{6.1} & \textbf{4.1} & \textbf{4.7} & \textbf{3.3} & \textbf{8.3} & \textbf{5.1} & \textbf{33.9} & 32.5 & \textbf{29.6} & \textbf{33.6} & \textbf{77.1} & \textbf{79.0} \\

     \midrule

     \multirow{6}*{\rotatebox[origin=c]{90}{{OF-9B}}} & \lightgray Clean & \multicolumn{2}{c|}{\lightgray 79.7} & \multicolumn{2}{c|}{\lightgray 60.1} & \multicolumn{2}{c|}{\lightgray 23.8} & \multicolumn{2}{c}{\lightgray 48.5} & \multicolumn{2}{c}{\lightgray 65.7} & \multicolumn{2}{c}{\lightgray 0.0}\\

     & MIX.Attack~\cite{tu2023many} & 45.9 & 25.4 & 33.7 & 18.0 & 13.4 & 8.8 & 39.8 & 36.0 & 59.0 & 53.3 & 31.6 & 49.2 \\
     
     & VT-Attack~\cite{wang2024break} & 49.7 & 31.6 & 38.6 & 23.0 & 16.1 & 13.2 & 39.7 & 39.5 & 65.1 & 62.7 & 25.0 & 37.9\\

     & AttackVLM-ii~\cite{zhao2023evaluating} & 27.7 & 10.1 & 20.1 & 9.3 & 10.0 & 7.4 & 37.8 & 33.3 & 53.3 & 48.1 & 46.1 & 59.8\\

     & VEAttack~\cite{mei2025veattack} & 7.5  & 3.7 & 8.7 & \textbf{3.2} & 12.5 & 5.7 & 34.0 & 32.8 & 60.6 & 59.6 & 52.3 & 61.5\\

     & PA-Attack (ours) & \textbf{4.7} & \textbf{3.6} & \textbf{5.4} & 3.7 & \textbf{6.6} & \textbf{4.8} & \textbf{33.0} & \textbf{32.6} & \textbf{47.6} & \textbf{45.9} & \textbf{63.4} & \textbf{66.4} \\

     \midrule

    \multirow{6}*{\rotatebox[origin=c]{90}{{LLaVa1.5-13B}}} & \lightgray Clean & \multicolumn{2}{c|}{\lightgray 119.2} & \multicolumn{2}{c|}{\lightgray 77.1} & \multicolumn{2}{c|}{\lightgray 39.0} & \multicolumn{2}{c}{\lightgray 75.6} & \multicolumn{2}{c}{\lightgray 84.1} & \multicolumn{2}{c}{\lightgray 0.0}\\

     & MIX.Attack~\cite{tu2023many} & 73.8 & 60.1 & 41.0 & 32.2 & 22.8 & 19.7 & 59.8 & 58.0 & 74.2 & 68.2 & 31.8 & 39.9 \\
     
     & VT-Attack~\cite{wang2024break} & 68.6 & 34.0 & 40.7 & 25.2 & 28.7 & 18.4 & 59.8 & \textbf{27.6} & 70.9 & 67.9 & 30.5 & 54.9\\

     & AttackVLM-ii~\cite{zhao2023evaluating} & 46.6 & 25.3 & 27.5 & 17.9 & 19.6 & 14.7 & 55.0 & 49.9 & 63.8 & 57.8 & 45.3 & 56.6\\

     & VEAttack~\cite{mei2025veattack} & 11.2  & 6.5 & 9.1 & 5.7 & 12.4 & 8.6 & 41.5 & 37.6 & 47.6 & 44.7 & 67.1 & 72.4 \\

     & PA-Attack (ours) & \textbf{5.4} & \textbf{3.9} & \textbf{4.1} & \textbf{2.0} & \textbf{7.0} & \textbf{6.5} & \textbf{36.5} & 34.4 & \textbf{31.4} & \textbf{27.9} & \textbf{77.3} & \textbf{79.8} \\
     
    \bottomrule
    
  \end{tabular}
  }

%% file: Tables/module.tex
\centering
\resizebox{\columnwidth}{!}{
    \scriptsize
\tabcolsep=0.12cm
  \begin{tabular}{cccccccc}
    \toprule
     PG & AE & TS & Step & COCO & Flickr30k & TextVQA & VQAv2 \\
     \midrule

      \lightgray \ & \lightgray \  & \lightgray \ & \lightgray 100 & \lightgray 93.8 & \lightgray 91.9 & \lightgray 72.8 & \lightgray 48.4\\
    
     $\checkmark$ &  &  & 100 & 95.5 & 94.8 & 76.3 & 52.2\\
      & $\checkmark$ &  & 100 & 93.7 & 92.6 & 72.3 & 50.9 \\
      $\checkmark$ & $\checkmark$ &  & 100 & 95.3 & 94.7 & 76.6 & 54.2\\

      \lightgray \ & \lightgray \  & \lightgray \ & \lightgray 150 & \lightgray 94.8 & \lightgray 93.7 & \lightgray 77.9 & \lightgray 48.8\\

      $\checkmark$ & $\checkmark$ &  & 150 & 96.2 & 95.5 & 81.1 & 54.4\\

      $\checkmark$ & $\checkmark$ & $\checkmark$ & 150 & \textbf{96.5} & \textbf{95.7} & \textbf{86.3} & \textbf{56.4}\\
      
    \bottomrule
  \end{tabular}
  }

%% file: Tables/lambda.tex
\centering
\resizebox{\columnwidth}{!}{
    \scriptsize
\tabcolsep=0.17cm
  \begin{tabular}{cccccc}
    \toprule
     $\lambda$ & COCO & Flickr30k & TextVQA & VQAv2 & Average \\
     \midrule

     0.5 & \textbf{96.6} & 95.1 & 83.8 & 51.3 & 81.7 \\

     1.0 & 96.5 & \textbf{95.7} & \textbf{86.3} & \textbf{56.4} & \textbf{83.7} \\

     2.0 & 95.5 & 94.7 & 84.1 & 55.8 & 82.5 \\
      
    \bottomrule
  \end{tabular}
  }

%% file: Tables/prototype.tex
\centering
\resizebox{\columnwidth}{!}{
    \scriptsize
\tabcolsep=0.15cm
  \begin{tabular}{ccccccc}
    \toprule
     $m$ & $w$ & $K$ & Flickr30k & TextVQA & VQAv2 & Average \\
     \midrule

     \multicolumn{3}{c}{\lightgray  baseline} & \lightgray 91.9 & \lightgray 72.8 & \lightgray 48.4 & \lightgray 71.0 \\
    
     1000 & 512 & 20 & 92.8 & 77.1 & 47.6 & 72.5\\
     2000 & 1024 & 20 & 92.6 & \textbf{79.8} & 47.5 & 73.3\\
     3000 & 1024 & 20 & \textbf{94.8} & 76.3 & \textbf{52.2} & \textbf{74.4}\\
     3000 & 1024 & 10 & 94.2 & 77.6 & 51.2 & 74.3\\
     3000 & 1024 & 30 & 92.3 &79.5 & 49.0 &73.6\\
     3000 & 2048 & 20 & 89.9 & 74.7 & 47.5 & 70.7\\
     % 4000 & 1024 & 20 & 91.2 & 77.1 & 46.3 & 71.5\\
      
    \bottomrule
  \end{tabular}
  }

%% file: Tables/distance.tex
\centering
\resizebox{\columnwidth}{!}{
    \scriptsize
\tabcolsep=0.17cm
  \begin{tabular}{ccccc}
    \toprule
     Guidance & COCO & Flickr30k  & VQAv2 & Average \\
     \midrule

     \lightgray baseline & \lightgray 93.8 & \lightgray 91.9 & \lightgray 48.4 & \lightgray 78.1 \\

    Mean sample & 95.1 & 94.1 & 56.2 & 81.8 \\

    Nearest sample & 64.6 & 86.2 & 33.4 & 61.4 \\

    Farthest sample & 96.3 & 95.2 & 50.2 & 80.6 \\

    Nearest prototype & 95.2 & \textbf{96.1} & 53.4 & 81.6 \\

    Farthest prototype & \textbf{96.5} & 95.7 & \textbf{56.4} & \textbf{82.9} \\
      
    \bottomrule
  \end{tabular}
  }

%% file: sec/X_suppl.tex
\clearpage
\setcounter{page}{1}
\maketitlesupplementary

\section{Out-of-distribution guidance dataset}

To investigate the sensitivity of PA-Attack to the distribution of the guidance dataset, we conducted additional experiments using \textbf{RVL-CDIP}~\cite{harley2015icdar} (document images) and \textbf{ScienceQA}~\cite{lu2022learn} (scientific diagrams) as guidance sources. These datasets represent a significant domain shift from the natural scenes typically found in COCO or Flickr30k. Specifically, we construct prototypes using 3,000 randomly sampled images from each dataset. As presented in Table~\ref{tab:guidance}, PA-Attack exhibits strong robustness to prototype sources, maintaining excellent performance even when the guidance data originates from vertical domains significantly different from the test images. 

\begin{table}[h]
\vspace{-0.1cm}
\renewcommand{\arraystretch}{0.92}
\caption{Attack performance with different guidance datasets.}
\vspace{-0.11cm}
\input{Tables/Guidance}
\label{tab:guidance}
\vspace{-0.1cm}
\end{table}

\section{More recent and diverse LVLMs}

To verify that our method remains effective against the latest advancements in multimodal learning, we extend our experiments to include \textbf{Qwen3-VL-8B}~\cite{Qwen3-VL} and \textbf{InternVL2-8B}~\cite{chen2024internvl, chen2024expanding} on RealWorldQA~\cite{realworldqa2024}, ODinW-13~\cite{li2022grounded} and POPE~\cite{li2023evaluating} datasets. For a fair comparison, we set baseline iterations $S=300$ and perturbation budget $\epsilon=8/255$, while configuring PA-Attack with $S_1=100, S_2=200$. As shown in Table~\ref{tab:qwen3}, our PA-Attack consistently demonstrates superior effectiveness across diverse LVLMs and tasks. 

\begin{table}[h]
\vspace{-0.1cm}
\renewcommand{\arraystretch}{0.95}
\caption{Attack performance on Qwen3-VL and InternVL2.}
\vspace{-0.11cm}
\input{Tables/Qwen_InternVL}
\label{tab:qwen3}
\vspace{-0.1cm}
\end{table}

\section{Larger perturbation budget}

While standard gray-box attacks typically operate under strict imperceptibility constraints (e.g., $\epsilon=2/255$ or $4/255$), it is crucial to evaluate attack scalability when these constraints are relaxed. We conducted additional experiments on LLaVA-1.5-7B with $\epsilon=8/255$. Results in Table~\ref{tab:eps8} demonstrate that PA-Attack maintains its superiority and scalability even under larger perturbation allowances.

\begin{table}[h]
\vspace{-0.13cm}
\renewcommand{\arraystretch}{0.95}
\caption{Attack performance with larger perturbation budget.}
\vspace{-0.11cm}
\input{Tables/eps8}
\label{tab:eps8}
\vspace{-0.14cm}
\end{table}

\section{Runtime comparison.}

\begin{figure}[h]
\centerline{\includegraphics[width=0.94\linewidth]{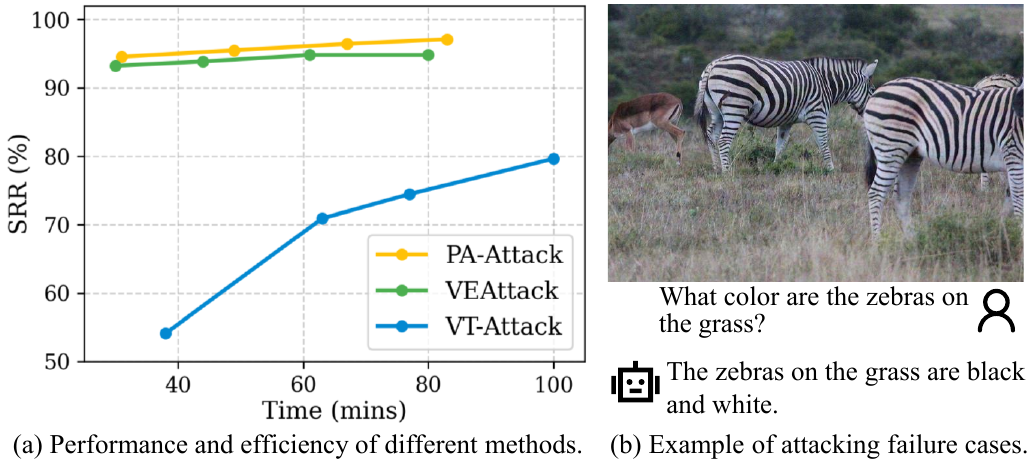}}
    \vspace{-0.5em}
	\caption{Ablation of time and an example of failure cases.}
    \label{f-time}
    \vspace{-0.5em}
\end{figure}

We analyze the trade-off between attack effectiveness (SRR) and computational cost (runtime) in Figure~\ref{f-time} (a). Results indicate that PA-Attack achieves superior SRR with only a marginal computational increase compared to VEAttack. Although our method introduces additional steps for prototype guidance and attention weight calculation, these operations are computationally lightweight compared to the gradient backpropagation through the vision encoder.

\section{Failure cases}

To understand the boundaries of our attack, we analyze specific failure cases. As illustrated in Figure~\ref{f-time} (b), attacks tend to be less effective when the textual query triggers strong inherent knowledge priors within the Large Language Model (LLM). In such instances, the LVLM relies more heavily on its pre-trained linguistic knowledge than on the visual context provided by the vision encoder. For example, when asked about the color of a zebra, the LLM's strong association between "zebra" and "black and white" may override the perturbed visual embeddings. This suggests that while PA-Attack effectively disrupts visual representations, downstream hallucinations or strong language priors can sometimes bypass vision-targeted perturbations.

\section{Effectiveness of Adversarial Training}

To further validate the robustness of our method, we evaluate PA-Attack against state-of-the-art vision encoder-specific adversarial training (AT) defenses, specifically TeCoA~\cite{mao2022understanding} and FARE~\cite{schlarmann2024robust}. The results in Table~\ref{tab:defense} show that while the performance of baseline methods like VEAttack degrades sharply under these defenses, PA-Attack consistently maintains a higher Score Reduction Rate (SRR). This resilience indicates that the adversarial features generated by our prototype guidance and attention refinement mechanisms are semantically more robust and harder to mitigate than simple gradient-based noise. PA-Attack proves to be a more challenging threat model for current defense strategies.

\begin{table}[h]
\vspace{-0.3cm}
\renewcommand{\arraystretch}{0.92}
\caption{Attack performance with different defense methods.}
\vspace{-0.31cm}
\input{Tables/defense}
\label{tab:defense}
\vspace{-0.28cm}
\end{table}

\section{More visualizations of predictions}
\label{sec:subjects}

We provide qualitative comparisons in Fig.~\ref{f-subject} of the paper and Fig.~\ref{f-subject1}, \ref{f-subject2}, and \ref{f-subject3} of the supplementary material to illustrate the different impacts of gray-box AttackVLM-ii, black-box M-Attack, and our gray-box PA-Attack. From a visual perspective, the perturbations generated by the black-box M-Attack are visibly more pronounced, while the gray-box methods generate much subtler and less perceptible noise, remaining visually closer to the original clean images.
Moreover, the output-level comparisons highlight the task generalization of our method. Both the gray-box AttackVLM-ii and the black-box M-Attack, which lack prototype-anchored guidance and attention enhancement, often fail to fully attack core attributes in the image captioning task. This partial success leads to a cascading failure, as the VQA and POPE tasks that query these specific, unchanged attributes are not successfully attacked. For instance, in Fig.~\ref{f-subject1}, both AttackVLM-ii and M-Attack still identify "a man is walking," which causes them to correctly answer the related VQA ("Is there someone crossing the street?") and POPE ("Is there a man in the street?") constructions.
In contrast, our PA-Attack completely changes all attributes, which could lead to more general attack successes in other tasks. In Fig.~\ref{f-subject1}, the subject is converted to "a clock", successfully deceiving all three tasks. These visualizations demonstrate that by leveraging prototype-anchored guidance and attention enhancement, PA-Attack achieves a more thorough and general attack by successfully altering core semantic attributes.

\section{More visualizations of attentions}
\label{sec:difference}

\subsection{Attention maps}

\begin{figure*}
\centerline{\includegraphics[width=1.0\linewidth]{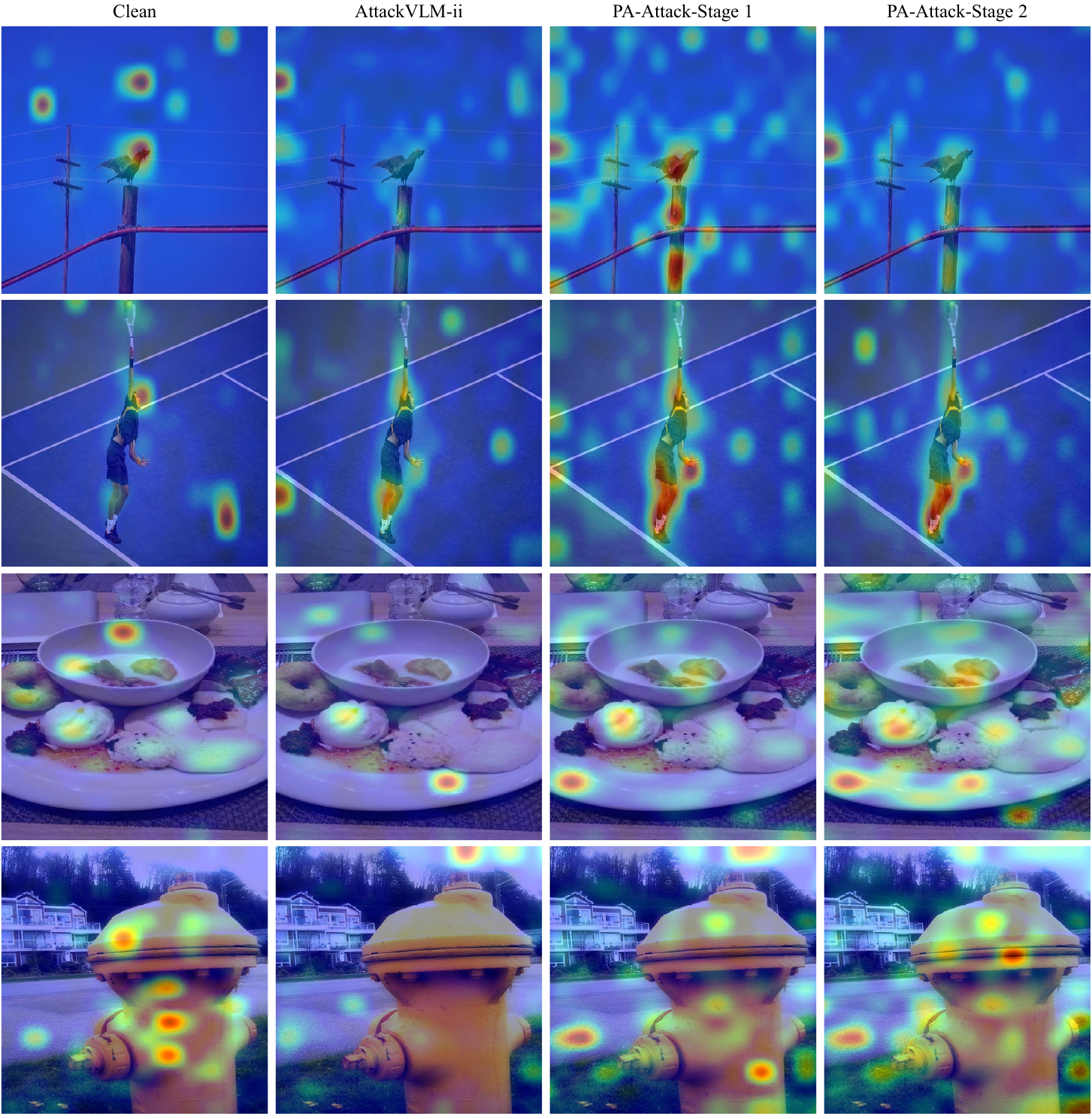}}
	\caption{\textbf{Comparison of the attention maps of clean images and those after different attacks.} The iterations of AttackVLM-ii are 100. We visualize the attention map of the first stage of our PA-Attack with 50 iterations and the second stage with 100 iterations. The areas that are more reddish have higher attention values.}
    \label{f-attnimages}
\end{figure*}

In Fig.~\ref{f-attnimages}, we provide a qualitative comparison by visualizing the attention maps of the vision encoder. We compare the attention on clean images against that on images perturbed by the AttackVLM-ii and both stages of our PA-Attack. In the "Clean" column, it can be seen that when processing clean images, the vision encoder distributes its focus across both the primary subject and the surrounding background, which allows it to extract comprehensive features necessary for downstream tasks. However, the attention maps for images in the second column attacked by AttackVLM-ii appear to force focus onto specific, high-vulnerability regions. For example, in the third row, it focuses on the plate itself, rather than the objects on it. PA-Attack demonstrates that the attention remains more generalized and semantically meaningful. In the third row, our method maintains a broader focus on both the plate and the diverse food items it contains. In addition, the attention enhancement module directs focus toward critical information. In the second row, PA-Attack achieves a much more comprehensive and complete attention coverage over the main subject. Finally, PA-Attack-Stage 1 in the third column and PA-Attack-Stage 2 in the fourth column reveal a shift in focus. This dynamic adjustment validates the effectiveness of our two-stage refinement, which could update and optimize the adversarial focus.

\subsection{Difference of attentions with iterations}

\begin{figure*}
\centerline{\includegraphics[width=1.0\linewidth]{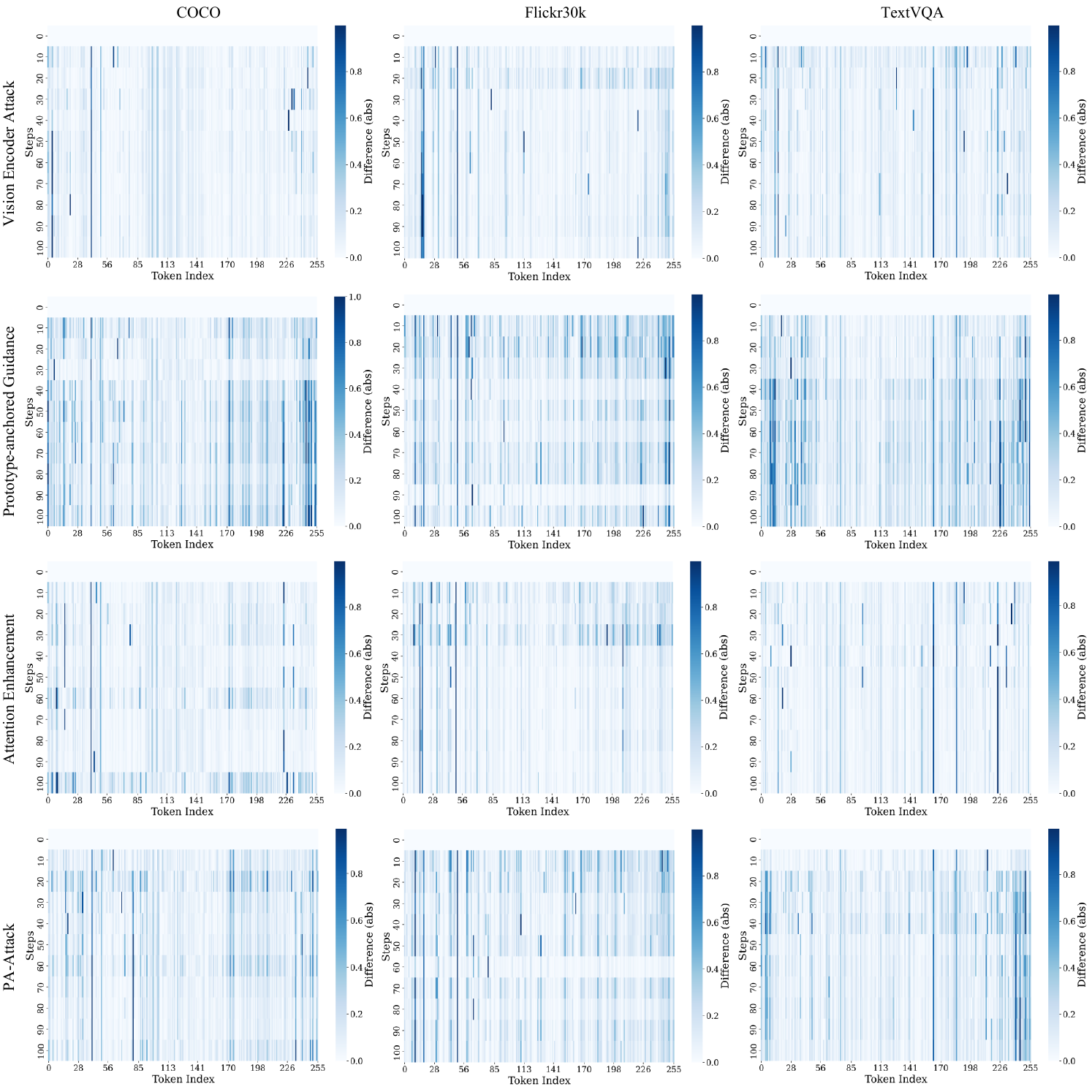}}
	\caption{\textbf{Evolution of token-wise attention values during the attack process.} The heatmaps visualize the deviation in token attention compared to the token of clean images with step 0. The x-axis represents token indices, and the y-axis represents the optimization steps. A darker shade indicates a greater difference in attention value from the clean image. The rows correspond to different modules, and the columns show results for three datasets.}
    \label{f-difference}
\end{figure*}

In Fig.~\ref{f-difference}, we visualize the evolution of token feature deviations during the optimization process on three datasets. The heatmaps illustrate the difference between token features at each attack iteration and the original clean features across all token indexes. Vision Encoder Attack consistently concentrates its perturbations on a few specific token indices throughout the attack. After introducing Prototype-anchored Guidance, influence is spread across a much broader range of tokens, indicating that the guidance prevents the optimization from collapsing onto a few specific features and promotes more attributes. Attention Enhancement adds a few more distinct vertical bands, representing its focus on specific important tokens. This reveals a key trade-off that prototype-anchored guidance succeeds in broadening the attack's scope but may reduce the relative focus on the most critical tokens. Our PA-Attack inherits the widespread, generalized deviation pattern from the prototype guidance, but it also exhibits stronger, more pronounced perturbations on the relatively important tokens identified by the attention enhancement module. This demonstrates that our method successfully achieves a trade-off, generalizing the attack vector while simultaneously prioritizing the most impactful tokens for optimization.

\begin{figure*}
\centerline{\includegraphics[width=1.0\linewidth]{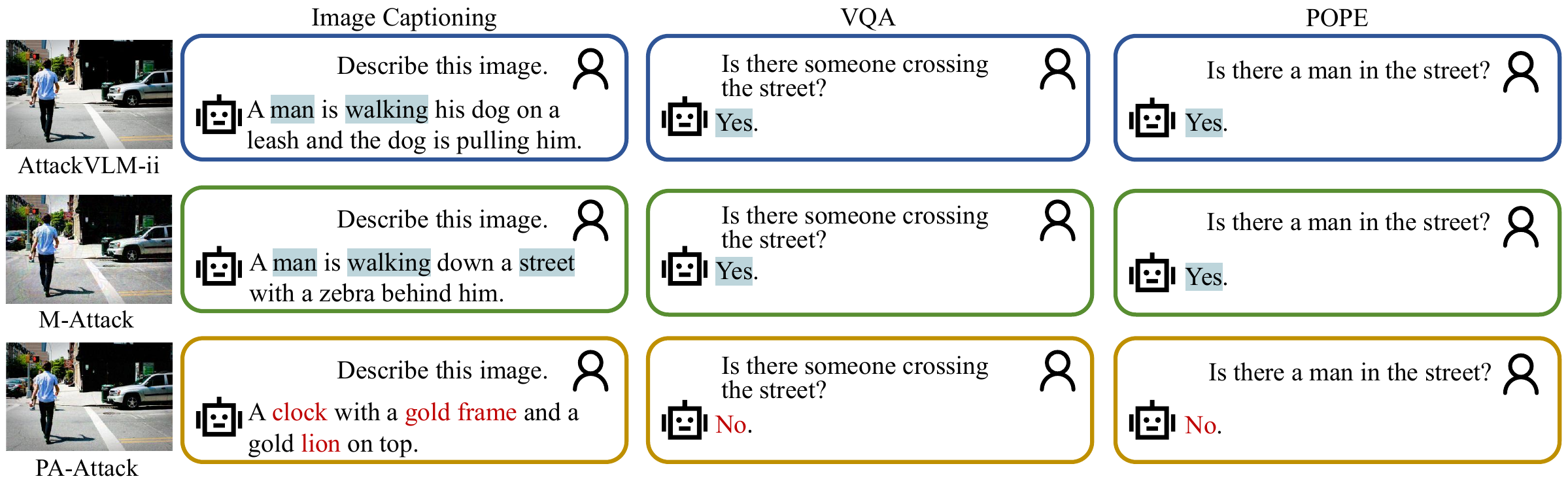}}
	\caption{\textbf{More comparison of the responses of LLaVa1.5-7B with different attacks.} The attributes with a blue background remain unchanged, while the red texts indicate that the attributes have changed.}
    \label{f-subject1}
\end{figure*}

\begin{figure*}
\centerline{\includegraphics[width=1.0\linewidth]{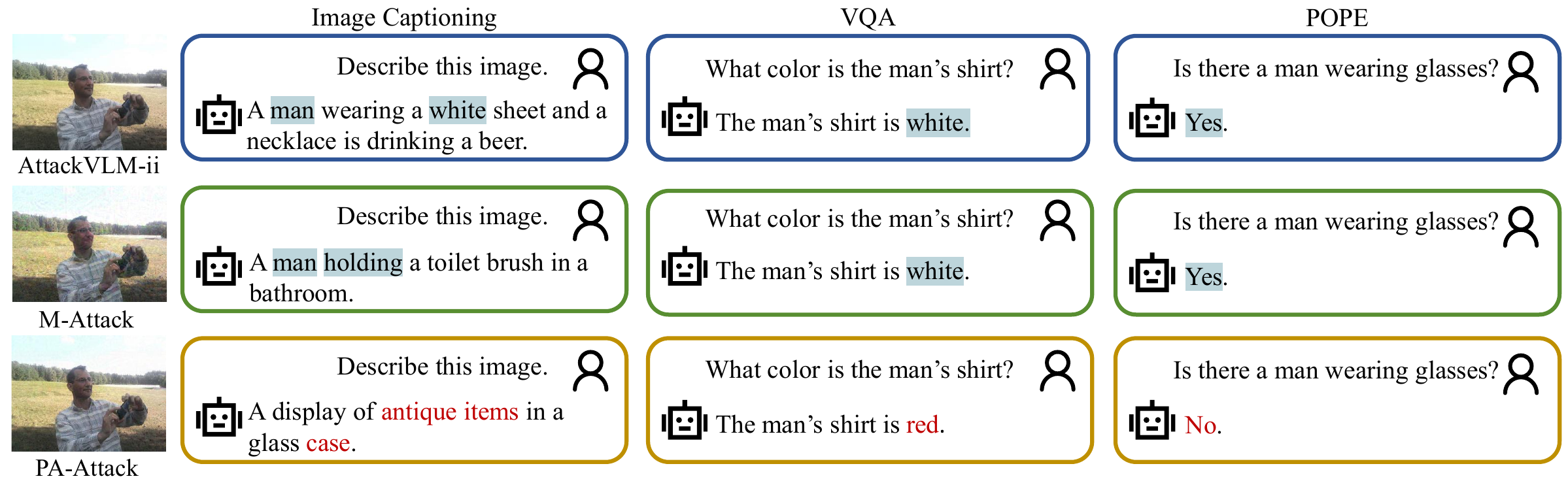}}
	\caption{\textbf{More comparison of the responses of LLaVa1.5-7B with different attacks.} The attributes with a blue background remain unchanged, while the red texts indicate that the attributes have changed.}
    \label{f-subject2}
\end{figure*}

\begin{figure*}
\centerline{\includegraphics[width=1.0\linewidth]{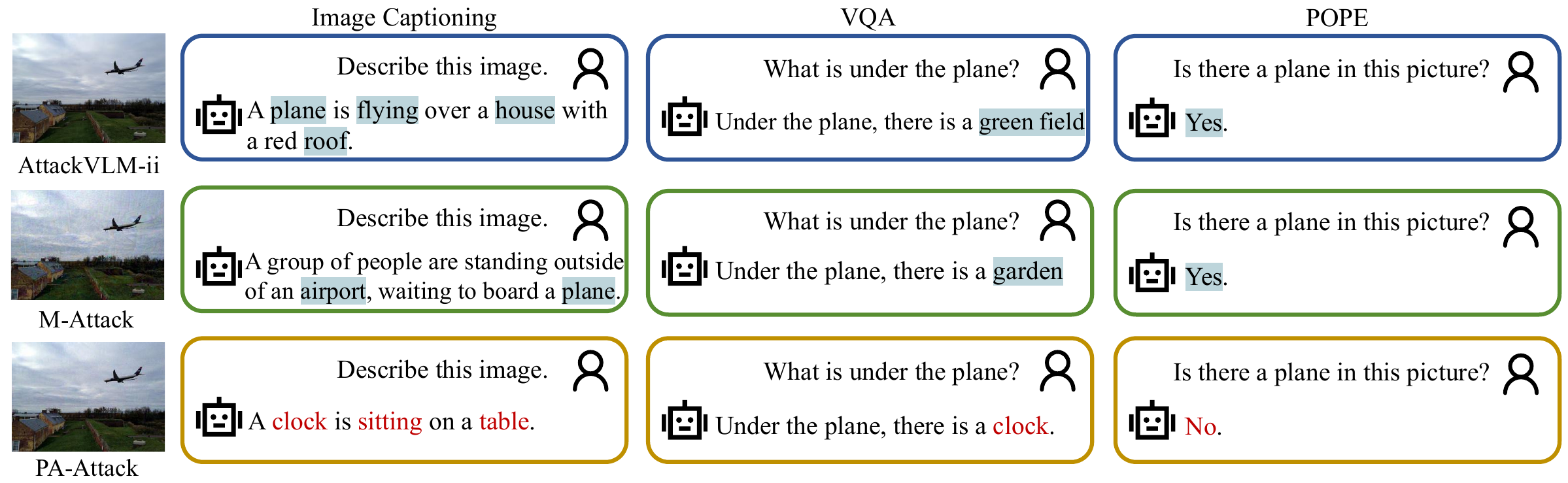}}
	\caption{\textbf{More comparison of the responses of LLaVa1.5-7B with different attacks.} The attributes with a blue background remain unchanged, while the red texts indicate that the attributes have changed.}
    \label{f-subject3}
\end{figure*}

%% file: Tables/Guidance.tex
\centering
\resizebox{0.97\columnwidth}{!}{
    \scriptsize
\tabcolsep=0.15cm
  \begin{tabular}{cccccc}
    \toprule
     Guidance & Distribution & COCO & Flickr30k & TextVQA & SRR $\uparrow$ \\
      \midrule

    \lightgray Clean & \lightgray In & \lightgray 115.5 & \lightgray 77.5 & \lightgray 37.1 & \lightgray 0.0 \\

     COCO & In & 4.1 & 3.3 & 5.1 & 92.8 \\

     RVL-CDIP & Out & 4.1 & 3.7 & 6.8 & 91.1 \\

     ScienceQA & Out & 4.2 & 3.6 & 5.8 & 92.0 \\
      
    \bottomrule
  \end{tabular}
  }

%% file: Tables/Qwen_InternVL.tex
\centering
\resizebox{\columnwidth}{!}{
    \scriptsize
\tabcolsep=0.15cm
  \begin{tabular}{cccccc}
    \toprule
     \multirow{2}*{Method} & \multicolumn{2}{c}{Qwen3-VL-8B} & \multicolumn{2}{c}{InternVL2-8B} & \multirow{2}*{SRR $\uparrow$} \\
      & RealWorldQA & ODinW-13 & RealWorldQA & POPE & \\
      \midrule

     \lightgray Clean & \lightgray 73.3 & \lightgray 35.5 & \lightgray 62.5 & \lightgray 88.6 & \lightgray 0.0 \\

     VT-Attack & 68.2 & 33.9 & 45.4 & 77.4 & 12.9 \\

     VEAttack & 58.2 & 14.4 & 46.4 & \textbf{63.8} & 33.4 \\

     PA-Attack & \textbf{57.0} & \textbf{9.8} & \textbf{43.4} & 64.8 & \textbf{38.0} \\
      
    \bottomrule
  \end{tabular}
  }

%% file: Tables/eps8.tex
\centering
\resizebox{0.95\columnwidth}{!}{
    \scriptsize
\tabcolsep=0.18cm
  \begin{tabular}{cccccc}
    \toprule
     Method & COCO & Flickr30k & TextVQA & VQAv2 & SRR $\uparrow$ \\
     \midrule

     \lightgray Clean & \lightgray 115.5 & \lightgray 77.5 & \lightgray 37.1 & \lightgray 74.5 & \lightgray 0.0 \\

     VT-Attack & 14.5 & 10.5 & 9.3 & \textbf{25.0} & 78.8 \\

     VEAttack & 5.3 & 4.0 & 6.5 & 36.3 & 81.0 \\

     PA-Attack & \textbf{3.6} & \textbf{1.9} & \textbf{4.4} & 31.9 & \textbf{84.9} \\
      
    \bottomrule
  \end{tabular}
  }

%% file: Tables/defense.tex
\centering
\resizebox{0.94\columnwidth}{!}{
    \scriptsize
\tabcolsep=0.17cm
  \begin{tabular}{ccccccc}
    \toprule
     \multirow{2}*{Method} & \multicolumn{2}{c}{COCO} & \multicolumn{2}{c}{Flickr30k} & \multicolumn{2}{c}{TextVQA} \\
      & TeCoA & FARE & TeCoA & FARE & TeCoA & FARE \\
      \midrule

     VEAttack & 43.9 & 51.1 & 24.2 & 34.1 & 16.5 & 22.2 \\

     PA-Attack & \textbf{25.8} & \textbf{47.5} & \textbf{15.1} & \textbf{30.7} & \textbf{10.6} & \textbf{14.1} \\
      
    \bottomrule
  \end{tabular}
  }